\documentclass[review,authoryear]{elsarticle}
\usepackage{booktabs}
\usepackage{array}
\usepackage{pdflscape}
\usepackage{amsmath}
\usepackage{amsfonts}
\usepackage{hyperref}
\usepackage{graphicx}
\graphicspath{{figures/}}
\usepackage{subfigure}

\usepackage{multirow}
\usepackage{rotating}
\usepackage[mathlines]{lineno}
\usepackage{color}
\usepackage{lineno,hyperref}
\modulolinenumbers[5]

\usepackage{graphics} 
\usepackage{graphicx}
\usepackage{epsfig} 
\usepackage{amsmath} 
\usepackage{amssymb}  
\usepackage{color}
\usepackage{booktabs}
\usepackage{caption}
\usepackage{verbatim}

\usepackage{float}

\journal{Unmanned Aerial Systems, Elsevier}

\begin{document}

\begin{frontmatter}

\title{RISCuer: A Reliable Multi-UAV Search and Rescue Testbed$^{\star}$}
\tnotetext[mytitlenote]{Supported by funding from King Abdullah University of Science \& Technology (KAUST)}

	\author[a]{Mohamed Abdelkader*}
		\cortext[corr-author]{Corresponding authors}
		\ead{mohamedashraf123@gmail.com}
	\author[b]{Usman A. Fiaz*}
		\ead{fiaz@umd.edu}
	\author[c]{Noureddine Toumi}
	\author[d]{\\Mohamed A. Mabrok}
	\author[a]{Jeff S. Shamma}

	\address[a]{Robotics Intelligent Systems \& Control (RISC) Lab, \\King Abdullah University of Science \& Technology, Thuwal, KSA}
	\address[b]{Department of Electrical \& Computer Engineering, Institute for Systems Research, University of Maryland, College Park, MD, USA}
	\address[c]{Group for Research in Decision Analysis, Polytechnique Montreal, Quebec, Canada}
	\address[d]{Department of Mathematics, School of Engineering, Australian College of Kuwait, Kuwait}
\begin{abstract}
We present the Robotics Intelligent Systems \& Control (RISC) Lab multiagent testbed for reliable search and rescue and aerial transport in outdoor environments. The system consists of a team of three multirotor unmanned aerial vehicles (UAVs), which are capable of autonomously searching, picking up, and transporting randomly distributed objects in an outdoor field. The method involves vision based object detection and localization, passive aerial grasping with our novel design, GPS based UAV navigation, and safe release of the objects at the drop zone. Our cooperative strategy ensures safe spatial separation between UAVs at all times and we prevent any conflicts at the drop zone using communication enabled consensus. All computation is performed onboard each UAV. We describe the complete software and hardware architecture for the system and demonstrate its reliable performance using comprehensive outdoor experiments, and by comparing our results with some recent, similar works. 
\end{abstract}
\vspace{2cm}
\begin{keyword}
Search\hfill and\hfill rescue,\hfill multiagent\hfill systems,\hfill unmanned aerial vehicles (UAVs), UAV testbeds, autonomous aerial grasping, reliable aerial transport
\end{keyword}

\end{frontmatter}


\begin{figure}[H]
\begin{center}
\includegraphics[width=12cm]{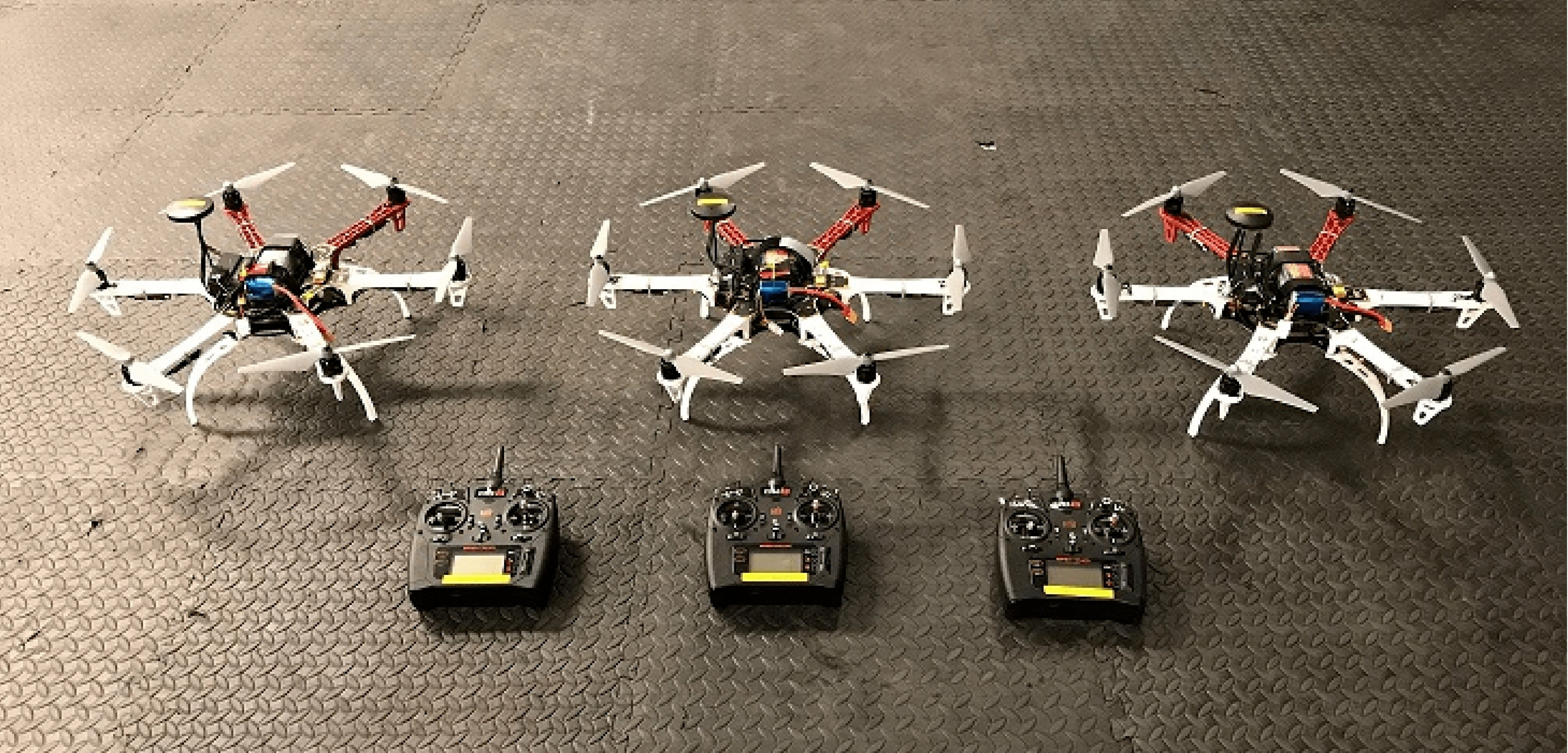}    
\caption{RISCuer: The RISC Lab cooperative multi-UAV testbed for search and rescue and autonomous aerial transport in outdoor environments.}
\label{fig:hexes}
\end{center}
\end{figure}

\section{Introduction}

Unmanned aerial vehicles (UAVs) find enormous utilization in several areas of interest to both academia and industry. Hence, there is a growing enthusiasm from scientists and engineers to push the operation and performance capabilities of these robots to their limit. Many of these efforts have resulted in significant advancements in airframe design, flight controls, reliable propulsion systems, and efficient power management for drones. UAVs serve as an ideal testbed for some of the recently proposed multiagent control algorithms \citep{mohammadi2020control}, \citep{fiaz2019hybrid}, \citep{abdelkader2017}, and are shown to have a major impact over many traditional industries as well. Examples include agriculture \citep{grenzdorffer2008}, \citep{zhang2012}, infrastructure monitoring \citep{adams2011}, \citep{ro2007}, public utility inspection \citep{agha2014}, and land surveying and construction \citep{d2012}. Thus, the significance of UAVs in modern industry cannot be overstated.

Despite this excess of existing literature in the area, it is quite noticeable however, that most of the existing implementations of multi-UAV systems are performed in indoor environments, i.e., in the presence of perfect positioning and precise localization, optimal lighting conditions, and a robust communication infrastructure. However, implementing a multi-UAV system is more challenging outdoors because of several external factors and disturbances in the environment. Therefore, in this chapter, we focus on the implementation and integration of a multi-UAV system (see Fig.~\ref{fig:hexes}), designed to complete a complex task cooperatively and autonomously in an outdoor environment. For our case study, we tackle the challenge of an outdoor multi-UAV search and rescue and autonomous aerial transport. Another constraint that greatly hinders the autonomous operation of UAVs outdoors is the need for onboard computation, because of the power and payload limitations on UAVs. In majority of the existing literature, the computation is performed off-board, which is acceptable for indoor lab experiments, but for realistic outdoor applications where a complete or substantially high degree of autonomy is desired, onboard computation requirement must be satisfied. Hence throughout this chapter, we only deal with and propose strategies which admit fully onboard control and computation capabilities for the UAVs involved. 

The rest of the chapter is organized as follows. Section~\ref{sec:survey} provides a brief literature survey on the existing state of the art for multiagent mission planning, aerial grasping, and search and rescue using UAVs. In Section~\ref{sec:problem}, we describe the problem and the underlying assumptions, and discuss our solution approach. In Section~\ref{sec:system}, we describe the complete system architecture and the various hardware/software components involved. Section~\ref{sec:fsm} demonstrates the finite state machine (FSM) for the mission. In Section~\ref{sec:vision}, we discuss strategies for object detection, localization and tracking using vision. Section~\ref{sec:grasping} details the aerial grasping mechanism, its actuation routine and our picking strategy for autonomous object transport. In Section~\ref{sec:comms}, we elaborate the communication framework for our multi-UAV system. Next, we demonstrate results from simulations and experiments in Section~\ref{sec:experiments}, and provide a quick comparison with some recent, similar works. Finally, we conclude with a brief discussion and some future directions in Section~\ref{sec:conclusion}.


\section{Related Work}\label{sec:survey}

There have been extensive efforts toward design enhancements, improved flight controls, and efficient path planning for UAVs over the past decade \citep{almurib2011}, \citep{lin2009}. Recent developments have encouraged roboticists to design and build UAVs that are capable of several useful operations which include but are not limited to, aerial grasping and transport \citep{pounds2011}, \citep{mellinger2013}, \citep{fiaz2017thesis}, collaborative construction using flying robots \citep{augugliaro2014}, \citep{6301066}, aerial perching on unstructured surfaces \citep{thomas2015}, and drone assisted search and rescue missions \citep{gholami2019drone} etc.

Many of these aforementioned applications typically require more than one robot in order to accomplish the task efficiently; for example consider the problem of aerial coverage \citep{yaziciouglu2013}. Clearly, it a multiagent distributed optimization problem that essentially desires multiple agents for communication-less coverage of a networked system \citep{yaziciouglu2017}. Again, we observe a lot of contributions have been made over the past decade in cooperative and collaborative implementations of UAVs for tasks such as simultaneous localization and mapping (SLAM) \citep{weiss2011}, vision-based autonomous UAV landing on moving platforms \citep{saripalli2002},\citep{beul2017fast}, and cooperative aerial transport of objects with multiple UAVs \citep{Michael2011}, \citep{nieuwenhuisen2017collaborative}.

It is evident from above, that aerial grasping is among the top research interests of people working in the field of aerial robotics. Besides, it can also be considered an integral component of UAV-based search and rescue missions \citep{fiaz2020fast}. Several useful techniques have been proposed for UAVs to grasp objects of various shapes, textures, weights, and sizes such as \citep{kessens2016}, \citep{hawkes2015}, and \citep{pounds2011}. All these works focus on the versatility of aerial grasping rather than its reliability and precision, which is indeed an interesting direction of research. However, for many practical and industrial applications, the need to grasp and transport objects reliably still remains a key objective. This is where ferrous aerial grasping comes to light. It is because of the well-known reliability and strength of the ferrous enclosures and their historic utilization in transportation of sensitive payloads and electronic components for several decades. In addition to the apparent physical protection, these enclosures also provide electromagnetic shielding to the transported payloads. Therefore, in this work, we specifically use our novel \citep{fiaz2019impulsive}, passive magnetic gripper design for the outdoor multiagent aerial transport. The mechanism uses the concept of passive aerial grasping of ferrous objects and enclosures \citep{fiaz2017passive}, combined with the dual impulsive release \citep{fiaz2018intelligent} of the payload at the drop zone. 

A bulk of recent multi-UAV search and rescue, cooperative aerial transport, and treasure hunt literature comes from Mohamed Bin Zayed International Robotics Challenge (MBZIRC) \citep{mbzirc}. This work is also motivated by the participation of Team KAUST at the inaugural version of MBZIRC, and is closely related to recent contributions from other participant teams such as \citep{nieuwenhuisen2017collaborative}, \citep{lee2019mission}, and \citep{beul2019team}. The key differences lie in our different approach to the mission, distinct system architecture, our novel and passive grasping mechanism, differences in actuation routine and communication protocols, and the mission execution itself. Therefore, throughout the rest of this chapter, we continue to highlight and compare these works with our method.


\section{Problem Description}\label{sec:problem}
We now describe the problem in detail along with the underlying assumptions. We then proceed with a summary of our approach for solving the problem.

As mentioned before, this work is motivated by one of the challenges posed in the inaugural version of MBZIRC. The problem setup considered in this work is as follows. A team of three UAVs has to collaborate in order to autonomously search, localize, track and pick up a set of static objects autonomously. The objects are known to be of ferrous material, and may consist of various sizes, shapes and colors which need to be transported to a dedicated single drop zone within limited time. The search area is an open outdoor space which is defined by a set of GPS coordinates which are known a priori. This problem brings up a set of practical research and system design questions regarding multi-UAV coordinated control, aerial grasping, and vision-based object detection and localization. 

\subsection{Assumptions}
\begin{figure}
\begin{center}
\includegraphics[width=8.4cm]{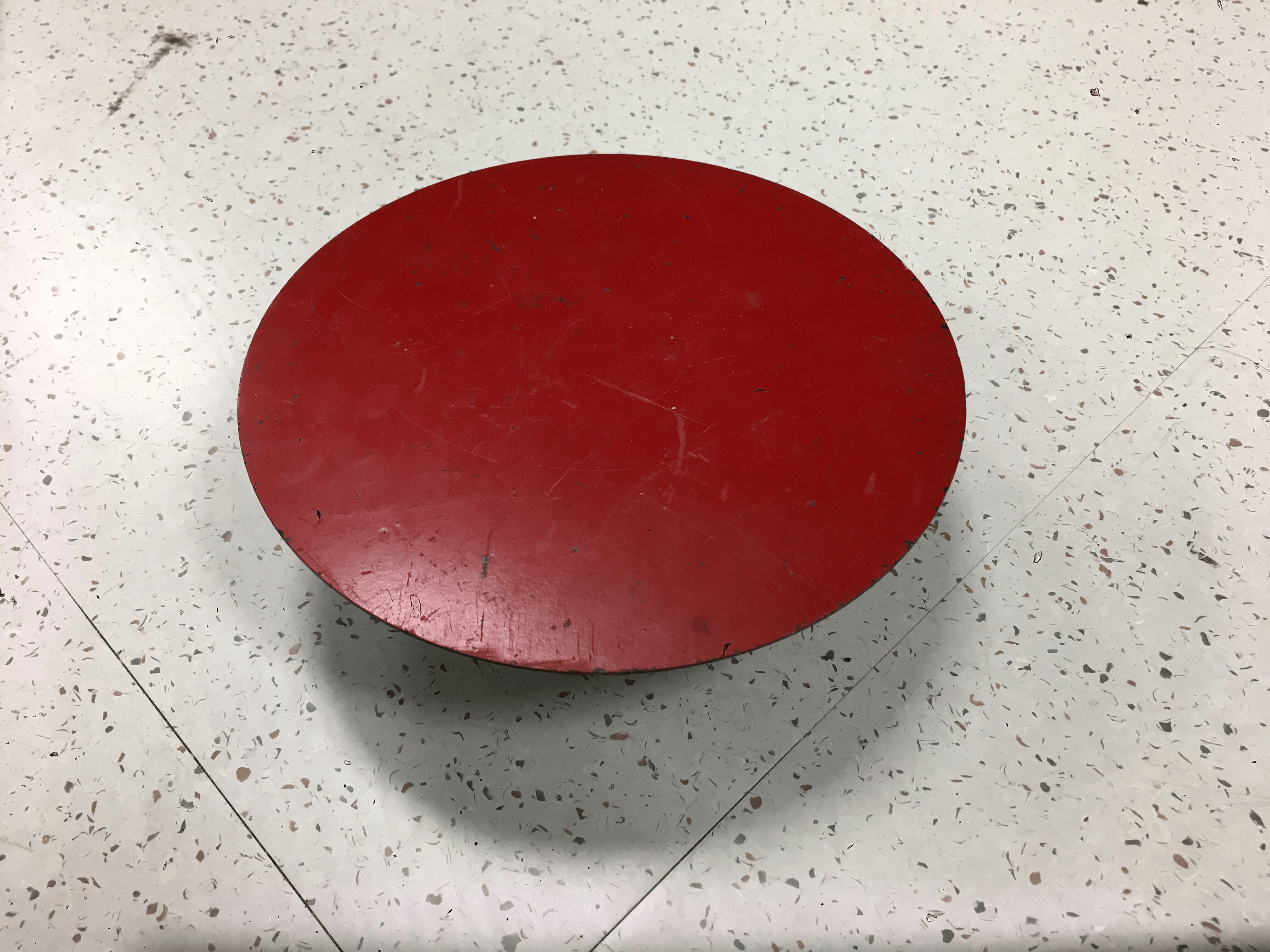}    
\caption{A sample payload used in this work. It is a 500 g disk of ferrous material with a diameter of 10 cm. } 
\label{fig:payload}
\end{center}
\end{figure}

Based on the problem statement, we consider the following assumptions.
\begin{itemize}
\item Each payload has a maximum weight of 500 g. This suffices to saying that a single UAV can pick up an object on its own and cooperative lifting of payloads is not necessary.
\item A dedicated wireless network is available (on demand) for each UAV to share information with each other as desired. In practice, a 2.4 GHz WiFi network is used for experiments.
\item All computation and decision making needs to be performed onboard each UAV; i.e., a centralized system is not allowed.
\item The top surface of the payloads is known to be flat. Furthermore to simplify the detection of objects, we assume the geometry of all objects to be circular. Thus, the payloads considered are circular colored ferrous disks (see Fig.~\ref{fig:payload}).
\item It is assumed that the search area has a rectangular geometry, with a known rectangular drop zone inside. This was specified in the MBZIRC challenge description as well.
\item The camera on each UAV is always facing downwards, i.e., a mechanical stabilization for the camera is present. This simplifies the problem of object localization using vision.
\end{itemize}

\subsection{Approach}
There are several ways to approach this challenging problem. One possible way could be to scan the whole search area for the objects using one or more UAVs. This will result in a map of the area with the detected object locations. One can then assign a given number of objects per UAV to transport them to the drop zone \citep{nieuwenhuisen2017collaborative}. It turns out, this approach is not the most efficient way of solving the problem though. A better approach could be to use partitioning of the workspace into several search areas and assigning a UAV to each of them separately \citep{beul2019team}. 

\begin{figure}[H]
\begin{center}
\includegraphics[width=8.4cm]{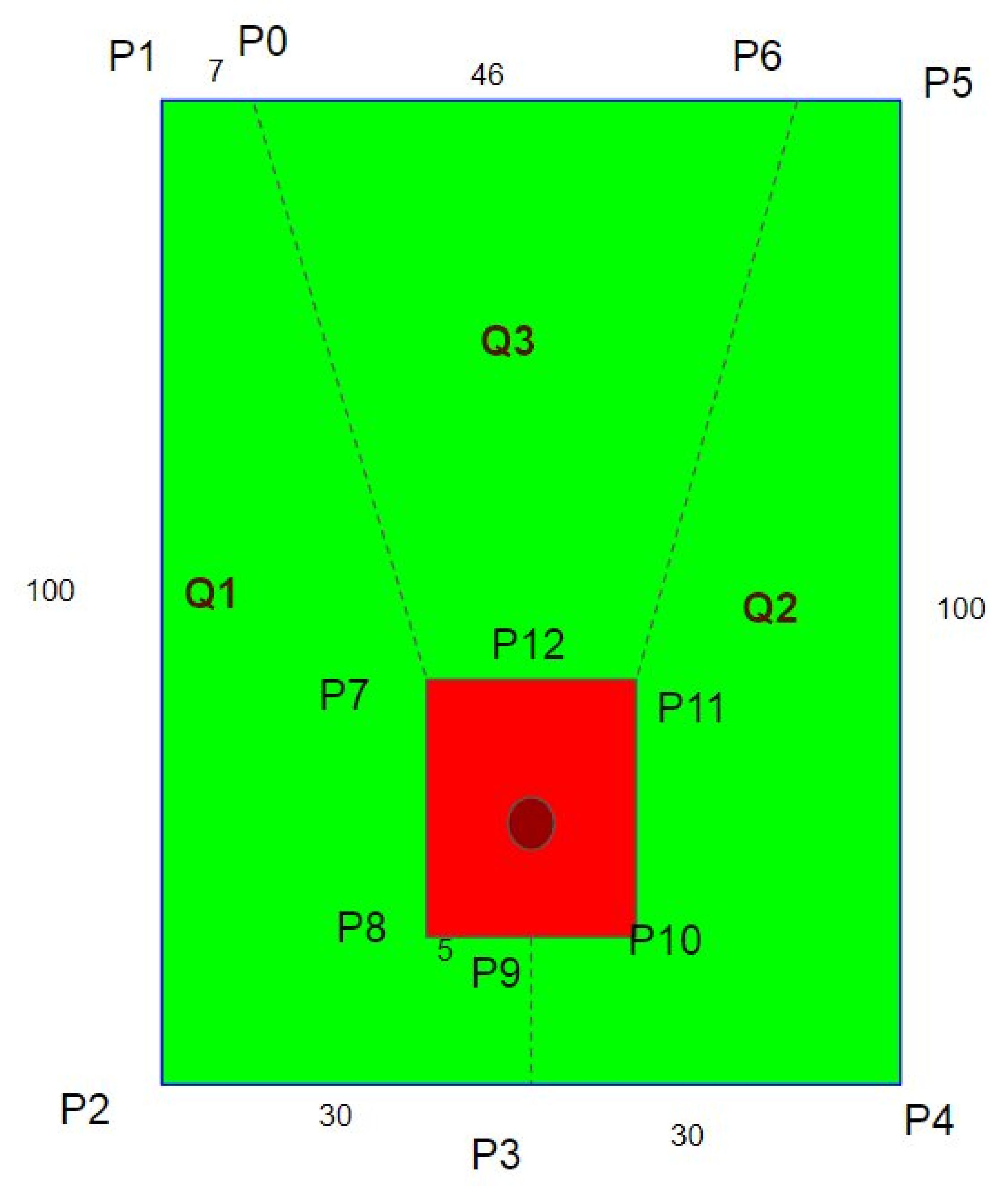}    
\caption{Partitioning of the rectangular field map. The three green partitions $Q1$, $Q2$, and $Q3$ represent the respective search areas for the three UAVs, while the red area represents the drop zone. Any conflicts at the drop zone are avoided using communication enabled consensus.} 
\label{fig:field_map}
\end{center}
\end{figure}

In this work, we use partitioning of the search area as well, to increase the speed of the search and rescue mission at hand. As shown in Fig.~\ref{fig:field_map}, we divide the workspace into three trapezoidal partitions of equal area. Each of the three UAVs is assigned to scan its respective partition for the objects. The scanning is performed in a uniform zig-zag fashion. Unlike \citep{lee2019mission}, as soon as a UAV detects an object, it proceeds to pick it up and transports it to the drop zone. After dropping the object, it returns to the same object location to restart its scanning routine. As is shown by simulation and experiments, this change enables our system to complete the mission faster than similar works, which also use partitioning methods. 

If a payload lies exactly on the boundary of two partitions, then a UAV which detects it first, has to pick it up. Further details on this partitioning, mission execution, and collision avoidance at the drop zone are provided in the following sections of the chapter. 

\section{System Architecture}\label{sec:system}
In this section, we provide a description of the hardware and software components used in the testbed.

\subsection{Hardware}
The testbed comprises of three identical hexarotors. Each of them is equipped with an autopilot for UAV control and navigation, a companion computer for high-level computation, a camera enabled vision system for object detection and localization, and communication system for information exchange between the UAVs and the ground control station (GCS) for monitoring.

\begin{figure}[H]
\begin{center}
\includegraphics[width=8.4cm]{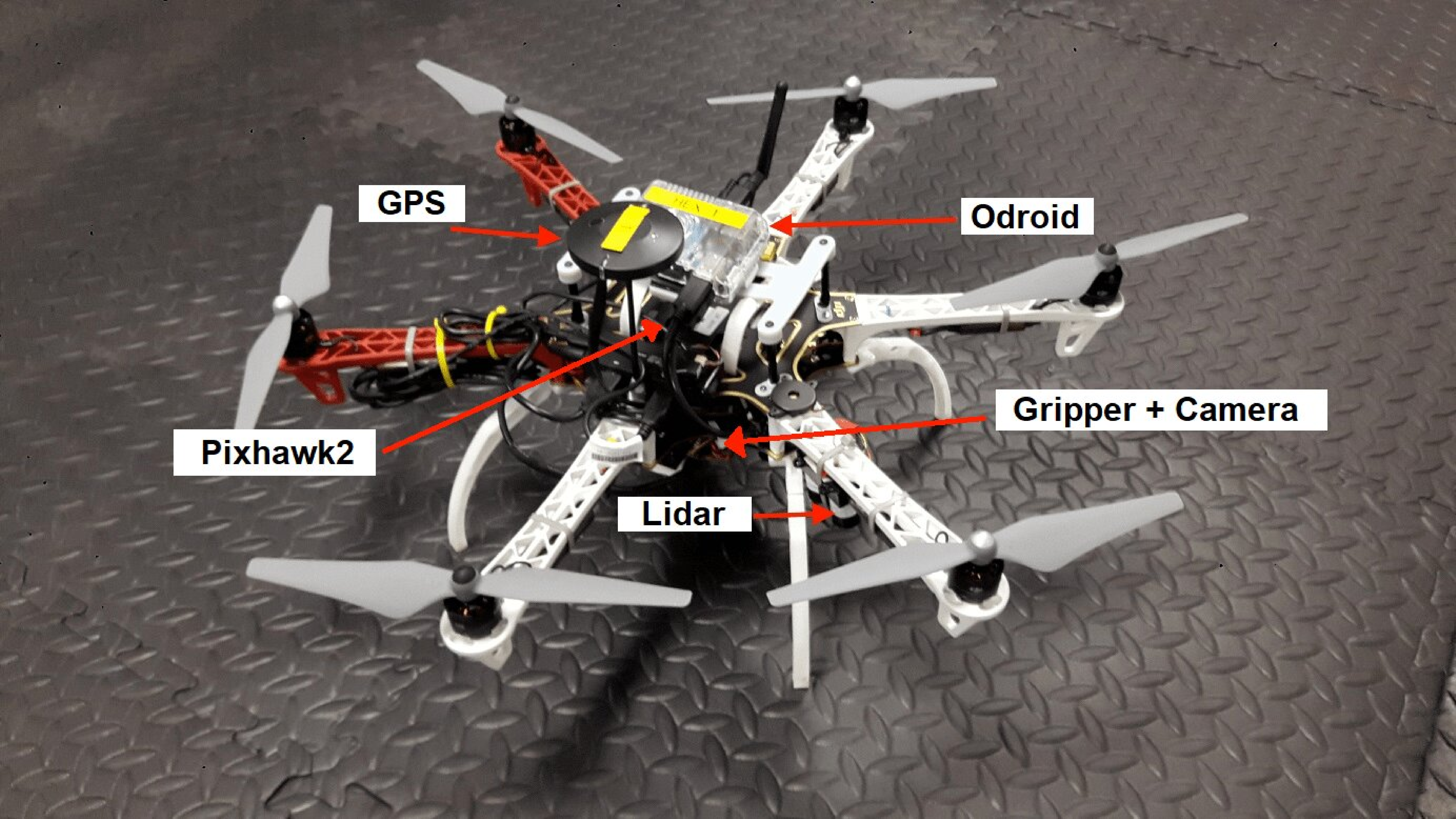}    
\caption{Fully equipped DJI F550 hexarotor platform.} 
\label{fig:hex_platform}
\end{center}
\end{figure}

\subsubsection{Hexarotor Platform}
Multirotor UAVs, e.g., hexarotors are known for their short flight time as compared to fixed-wing and other vertical take-off and landing (VTOL) platforms. That is because multirotors rely heavily on the thrust generated by their power hungry propulsion systems to stay airborne. However, multirotors have more agility and can hover in place, a trait which fixed-wing UAVs cannot generally achieve. We use an off-the-shelf hexarotor frame, the DJI Flamewheel F550, with customized onboard components (see Fig. \ref{fig:hex_platform}). Although we have tested a good number of quadrotor platforms as well, we decided to work with hexarotors as they provide more stability, agility, and an adequate payload capacity with a decent flight time of 20 minutes for the mission. The propulsion system, DJI E310 was selected because it provides enough thrust to carry a maximum payload of $2.5$ kg. A list of the main UAV components is given in Table \ref{tab:uav_components}.

\begin{table}[t]
\caption{UAV Hardware Components}
\centering\small
\begin{tabular*}{\linewidth}{@{\extracolsep{\fill}}p{0.3\linewidth}p{0.65\linewidth}@{}}
\toprule
\textbf{Item} & \textbf{Description}\\
\midrule
Frame: & DJI flamewheel F550 hexacopter\\
Propulsion system: & DJI E310 $900KV$ with $9$ inch propellers\\
Battery: & 10Ah 4S LiPo battery\\
Flight controller: & Pixhawk 2 (the cube)\\
On-board computer: & Odroid XU4\\
Altitude sensor: & LiDAR Lite v3 sensor with $40$m range\\
Camera: & ELP fish-eye camera\\
Gripper: & Custom passive design\\
\bottomrule
\end{tabular*}
\label{tab:uav_components}
\end{table}

\subsubsection{Autopilot}
\begin{figure}[]
\begin{center}
\includegraphics[width = 4cm]{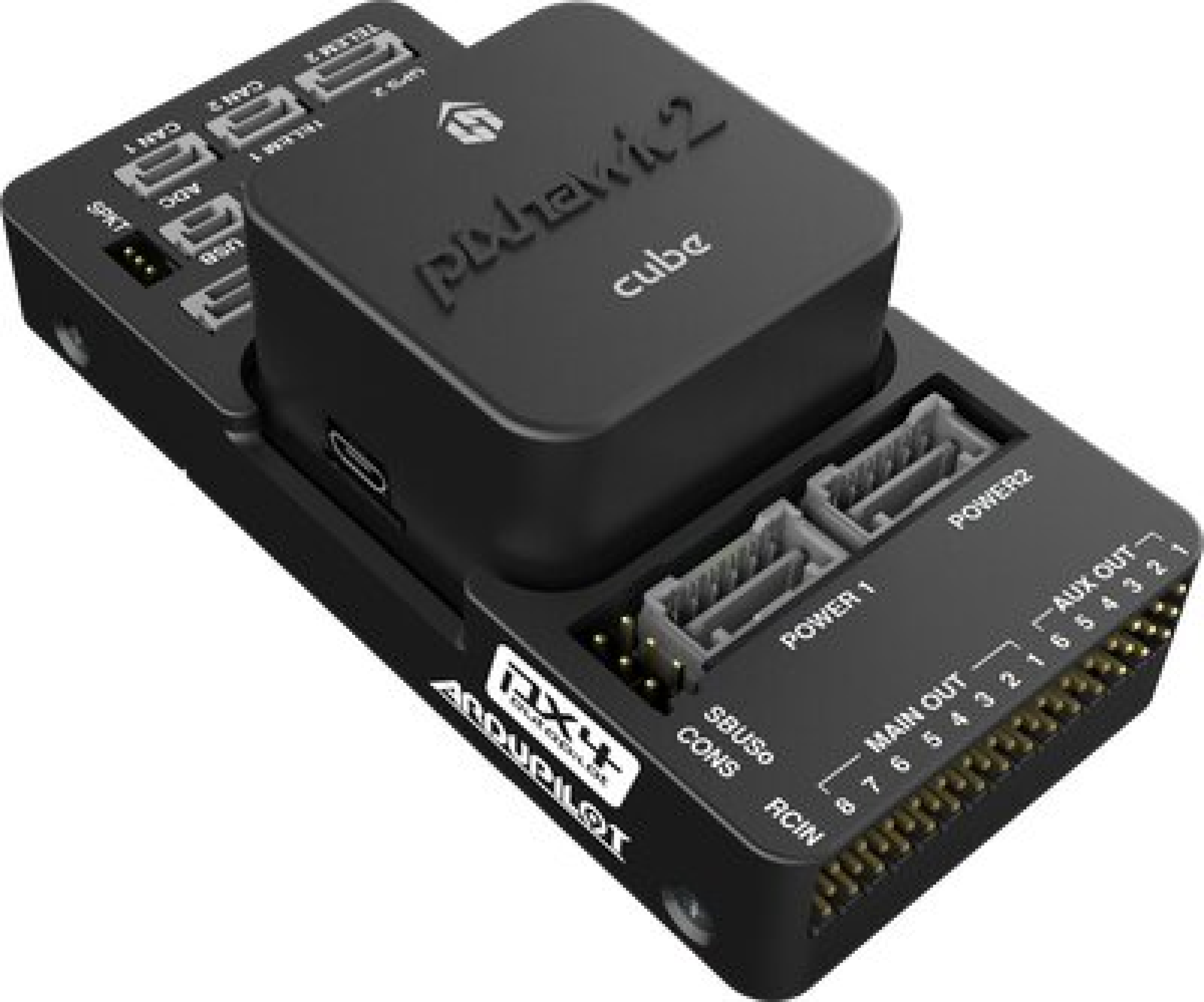}    
\caption{Autopilot: Pixhawk2 flight controller.} 
\label{fig:pixhawk2}
\end{center}
\end{figure}

We use the open-source Pixhawk2 flight controller (see Fig.~\ref{fig:pixhawk2}) along with the PX4 autopilot firmware for autonomous control and navigation of the UAVs. The PX4 software also allows us to use a companion computer, which is used to perform high-level algorithmic computations; for example vision processing, to send high-level commands such as attitude, velocity, and position set-points which the autopilot can then track. This control scheme allows the companion computer to focus on mission planning by leaving the low-level control load to the PX4 autopilot.

\subsubsection{Companion Computer}
\begin{figure}
\begin{center}
\includegraphics[width = 4cm]{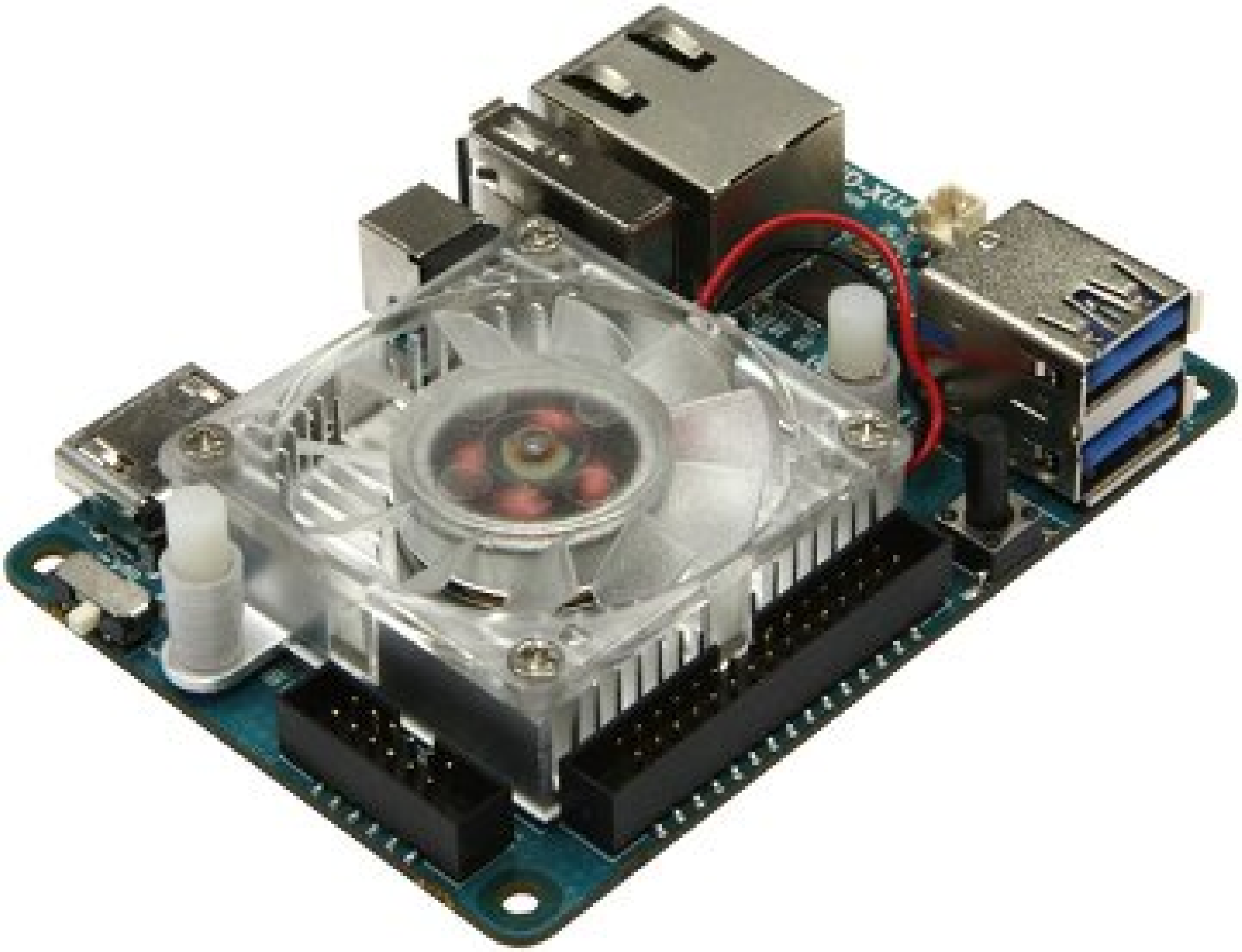}    
\caption{Odroid XU4: Onboard companion computer.} 
\label{fig:odroid}
\end{center}
\end{figure}

A companion computer is an embedded low power computing module that usually runs a version of Linux OS onboard a UAV. In our system, we use an Odroid XU4 (see Fig.~\ref{fig:odroid}) to: (1) execute onboard vision algorithms for object detection and localization, and (2) to execute the state machine which manages the overall system transitions. The Odroid board weighs around 70 g and is powered by a regulated 5V supply from the main battery.

\subsubsection{Sensors}
\begin{figure}[]
\begin{center}
\includegraphics[width = 8.4cm]{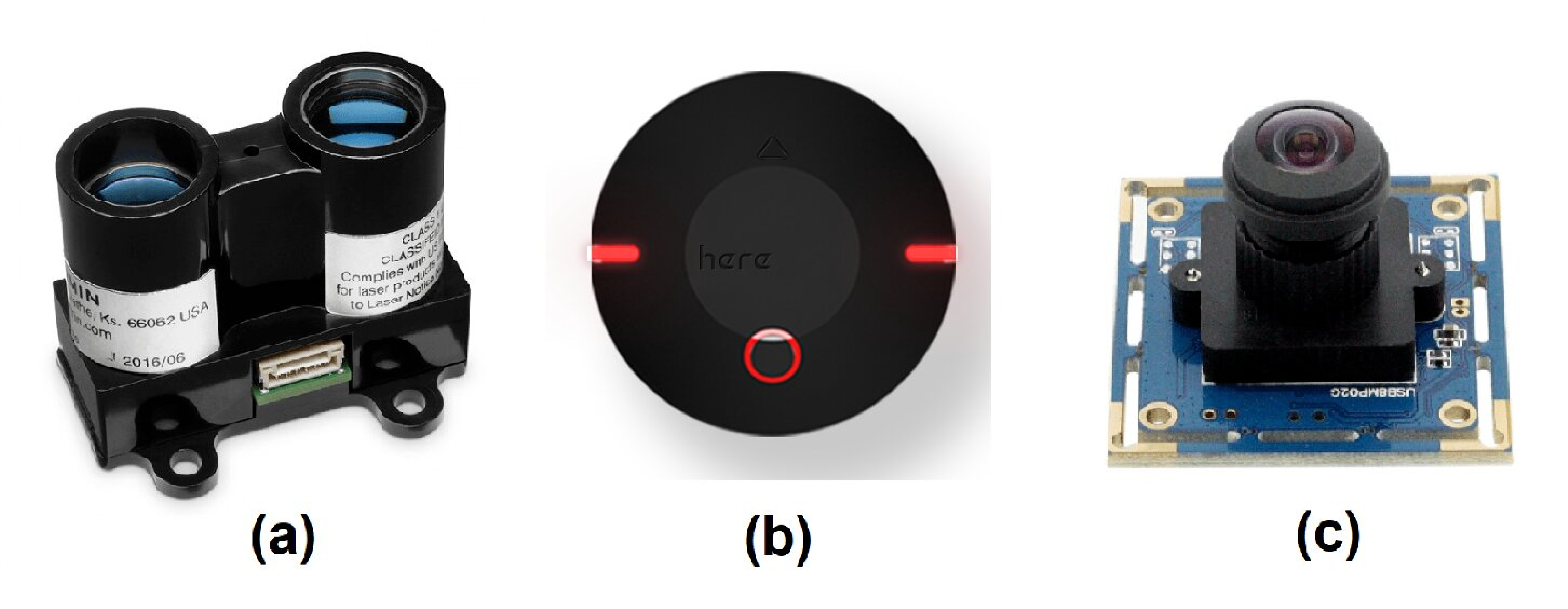}    
\caption{Onboard sensors: (a) LiDAR Lite altitude sensor, (b) Here+ GPS receiver, and (c) ELP fish-eye camera module.} 
\label{fig:sensors}
\end{center}
\end{figure}

In addition to the inertial measurement unit (IMU) which is embedded in the flight controller for attitude stabilization, we use the following three main sensors for localization and object detection (see Fig.~\ref{fig:sensors}):
\begin{itemize}
\item LiDAR Lite v3: A distance sensor which provides a much more precise altitude estimate than barometer-based altitude sensor; this allows us a precise altitude control at low altitudes during object picking.
\item Here+ GPS receiver: We used this model as it provides more accurate global positioning accuracy compared to many other products which we tested before.
\item 170-degree FoV fish-eye camera: An ELP wide angle camera; it helps in object detection at low altitudes for accurate aerial grasping. The camera is mounted on to a customized ultra-nano stabilization gimbal to provide a horizontal image capture, which makes the object localization process much easier.
\end{itemize}

\subsubsection{The Gripper}
A customized gripper is designed to grasp ferrous objects with a reliable pick up and drop confirmation message using our novel design \citep{fiaz2019impulsive}. This feedback information is critical for autonomy of the aerial grasping operation. The gripper uses a specific configuration of permanent magnets embedded with a proximity sensor for grasping, and a dual impulsive release mechanism for drop. The utilization of permanent magnets gives our design numerous advantages over other grasping techniques discussed in Section~\ref{sec:survey}. We cover the essentials of aerial grasping and release mechanism in Section~\ref{sec:grasping}. Further details on the gripper design can be found in our previous work \citep{fiaz2018intelligent}.


\subsection{Software}
The system software is distributed over two main components. The first component is the flight controller which receives set-point commands from the onboard computer, which is the second component. The onboard computer runs a state machine which manages the drone strategies starting from takeoff until the end of the mission. It also receives image frames from a USB wide-angle camera, and then runs an OpenCV-based vision algorithm which detects closest objects and converts the locations in image frames to relative position estimates. Finally, the velocity set-points are generated and sent to the flight controller to guide the drone for object search, picking, or dropping.

\begin{figure}
\begin{center}
\includegraphics[width = 11cm]{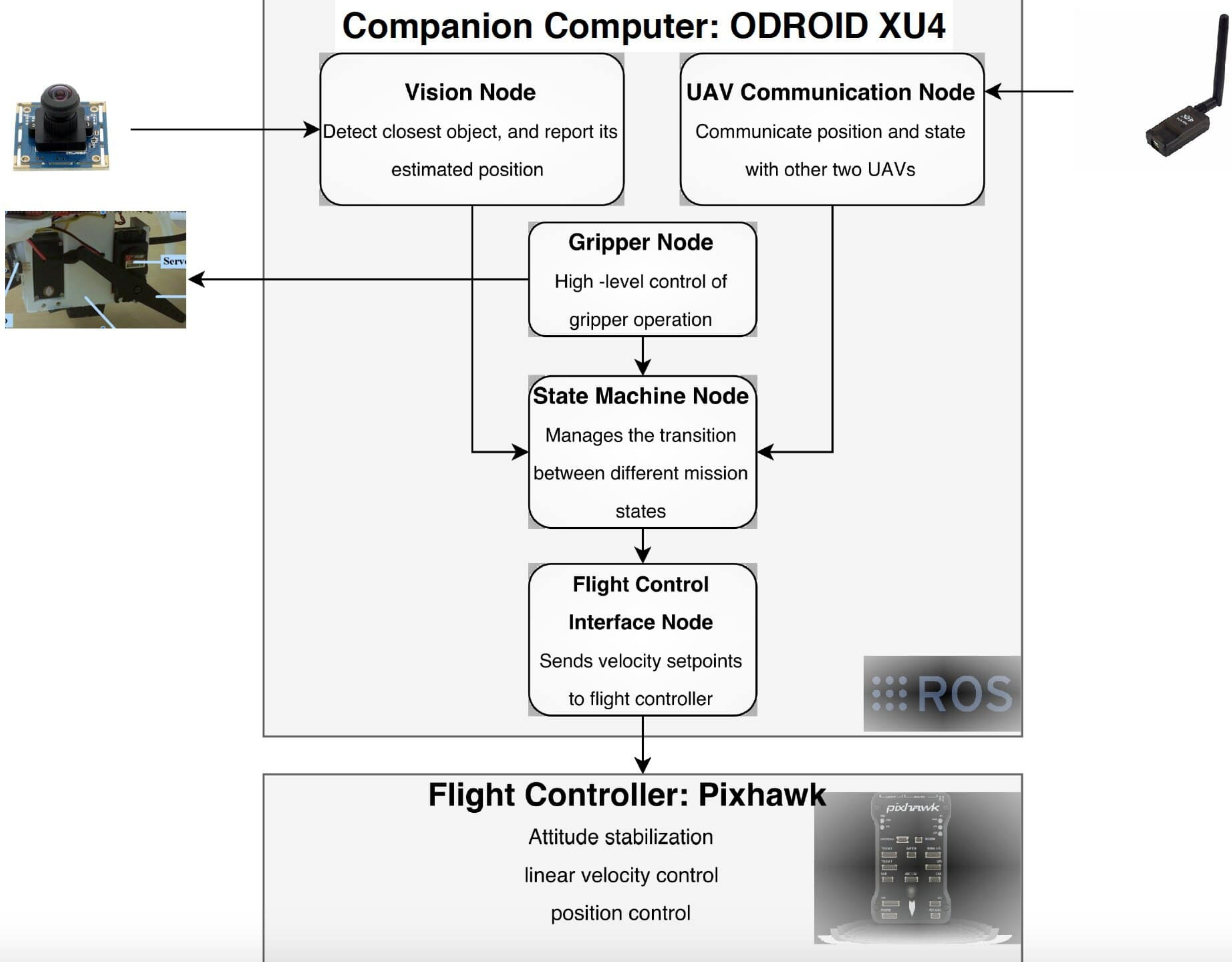}    
\caption{Software components of the system are distributed over two main parts: (1) A dedicated flight controller that handles real-time low-level vehicle stabilization and command tracking, and (2) a high-level companion computer which executes the remaining mission planning software.} 
\label{fig:ch3_diagram}
\end{center}
\end{figure}

The onboard computer software runs in Ubuntu Linux operating system, and we use the robot operating system (ROS)\footnote{http://www.ros.org} to conveniently interface the different software components. Figure~\ref{fig:ch3_diagram} shows the software architecture for each of the three UAVs in the system.

\section{State Machine Description}\label{sec:fsm}
\begin{figure}
\begin{center}
\includegraphics[width=10cm]{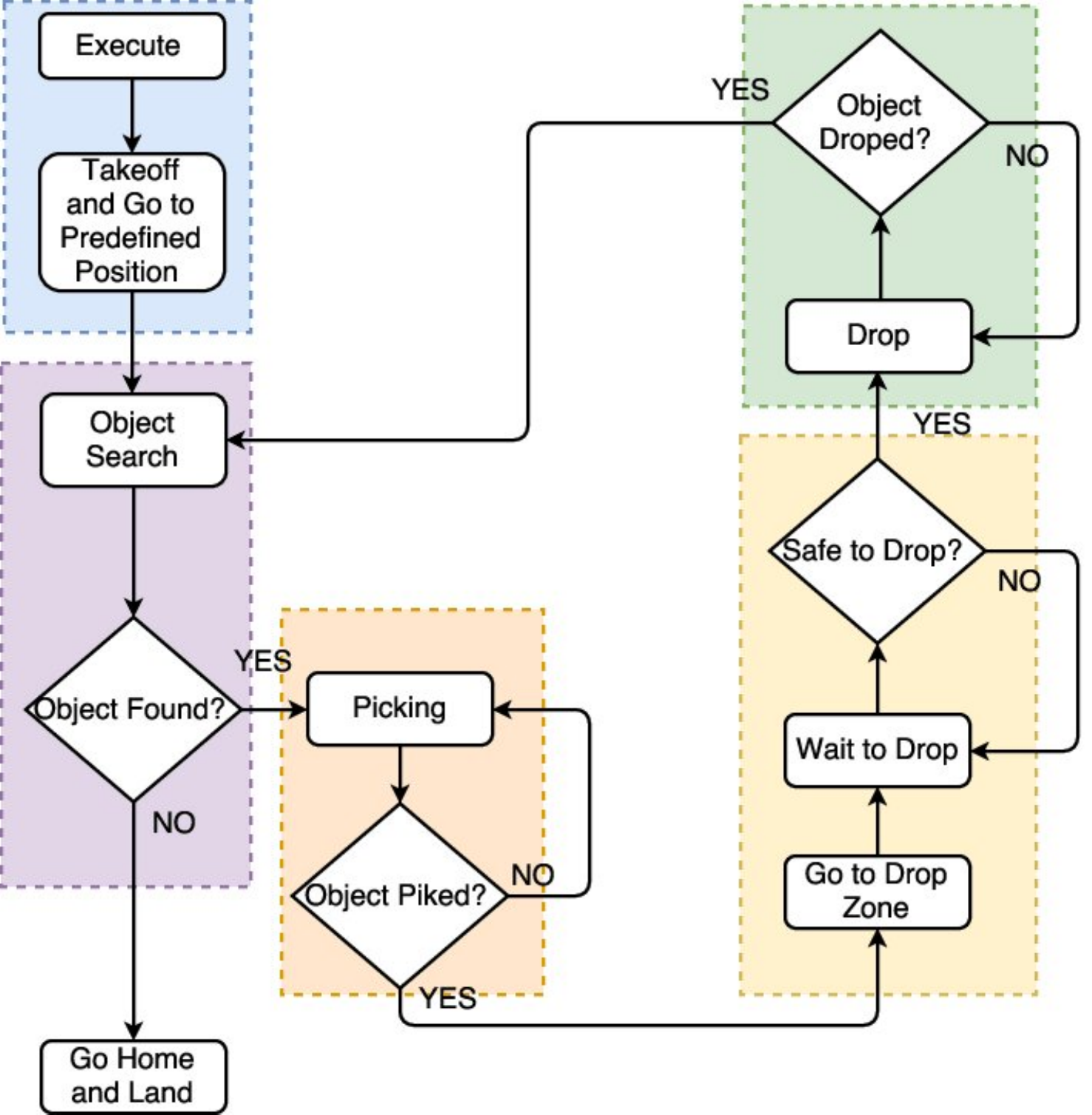}    
\caption{Flow diagram of the state machine for the mission.} 
\label{fig:SM_CH3}
\end{center}
\end{figure}

A finite state machine (FSM) is required in order to manage autonomous transitions of the system during the mission, from auto-takeoff, object search and transportation, to landing. The flow diagram of the FSM is shown in Fig.~\ref{fig:SM_CH3}. Now, we provide a brief description of each of the states of the FSM.

\subsection{Takeoff and Go to a Predefined Position}
This is an initialization state, where UAVs go to a predefined start location in their assigned operational area or partition.

\subsection{Object Search }
Once each UAV arrives at the predefined initial position, it automatically switches to \textit{Object Search} state. In this state, each UAV scans its own assigned area looking for objects. The scanning trajectories are designed to allow maximum distance between the UAVs to avoid collisions during the object search phase. If an object is detected, the state machine switches to the \textit{Object Picking} state.

\subsection{Object Picking}
In case an object is detected, the UAV will switch to \textit{Object Picking} state. It will keep trying to pick the object until it succeeds to do so. In each trial, it will descend gradually and check whether the object is well placed for picking. Otherwise it will ascend gradually to get more field of view. These steps are repeated until the UAV succeeds to pick up the object. If the object has been successfully picked, a sensor attached to the gripper will be activated and so the drone will switch to the \textit{Go to drop} state. If picking is not successful, it will switch back to \textit{Object Search}.

\subsection{Go to Drop}
Once an object is collected, the drone switches to the \textit{Go to Drop} state, in which it goes to a predefined spot around the perimeter of the drop zone. Then, it starts communicating with other UAVs to negotiate its eligibility to enter the drop zone, and this is done in the next state.

\subsection{Waiting to Drop}
For each UAV, there is a pre-assigned waiting spot where it must wait until there is no other UAV operating inside the drop zone. This state is the only state that requires communication between the agents and in case there were two agents waiting for access permission, the permission is granted according to a priority policy i.e., first come first serve. This simple yet effective strategy ensures that all agents will operate without any risk of collision.

\subsection{Drop}
In case none of the agents is inside the drop zone, the drone navigates to the drop spot inside. Once the drop spot is reached, the drone sends a command to the gripper to release the object. Sensors on the gripper send a feedback signal to confirm whether the drop was successful. If the operation is indeed successful, the drone switches to the \textit{Object Search} state.

\subsection{Go Home and Land}
After scanning the whole area and not finding any new object, the UAV will then switch from the \textit{Object Search} state to \textit{Go Home and Land} state, during which it flies towards a position called home spot where it lands. By doing so, the mission can be declared as accomplished for this particular UAV.


\section{Object Detection and Localization}\label{sec:vision}

Object localization is an essential step to guide the UAV to an accurate picking spot. For an object to be localized, it first needs to be detected. A monocular camera is used along with blob detection algorithm \citep{blob}, to detect objects of specific color and report their image pixel coordinates with respect to the image frame. If more than one object is detected, then the closest object is selected. In order to know how close the object is to the UAV, we use an empirical model which fuses the UAV altitude from ground with the reported object pixels in the image, to provide an accurate estimate of the object location with respect to the UAV. Such a model can be obtained by camera calibration process at a specific altitude. In this section, we explain the camera calibration process and the UAV-to-object control set-point calculations.

\subsection{Camera Calibration}

There have been several works related to aerial object tracking using different methodologies depending on the mission requirements and available tools. In particular, vision-based aerial object tracking has been an active field of research in computer vision community over the past decade \citep{1657153}, \citep{7531961}, \citep{2017arXiv170306527W}.

We use a fusion of vision-based object detection and UAV altitude information to accurately localize colored ground objects relative to the UAV coordinate frame. The approach mainly relies on an empirical model based on camera calibration with respect to a certain fixed altitude. The empirical model takes as inputs: $(1)$ The pixel coordinates of the center of the detected object and $(2)$ the current altitude of the UAV, and outputs a position estimate of the coordinates of the object relative to the UAV coordinate frame.

\begin{figure}
\begin{center}
\includegraphics[width=10cm]{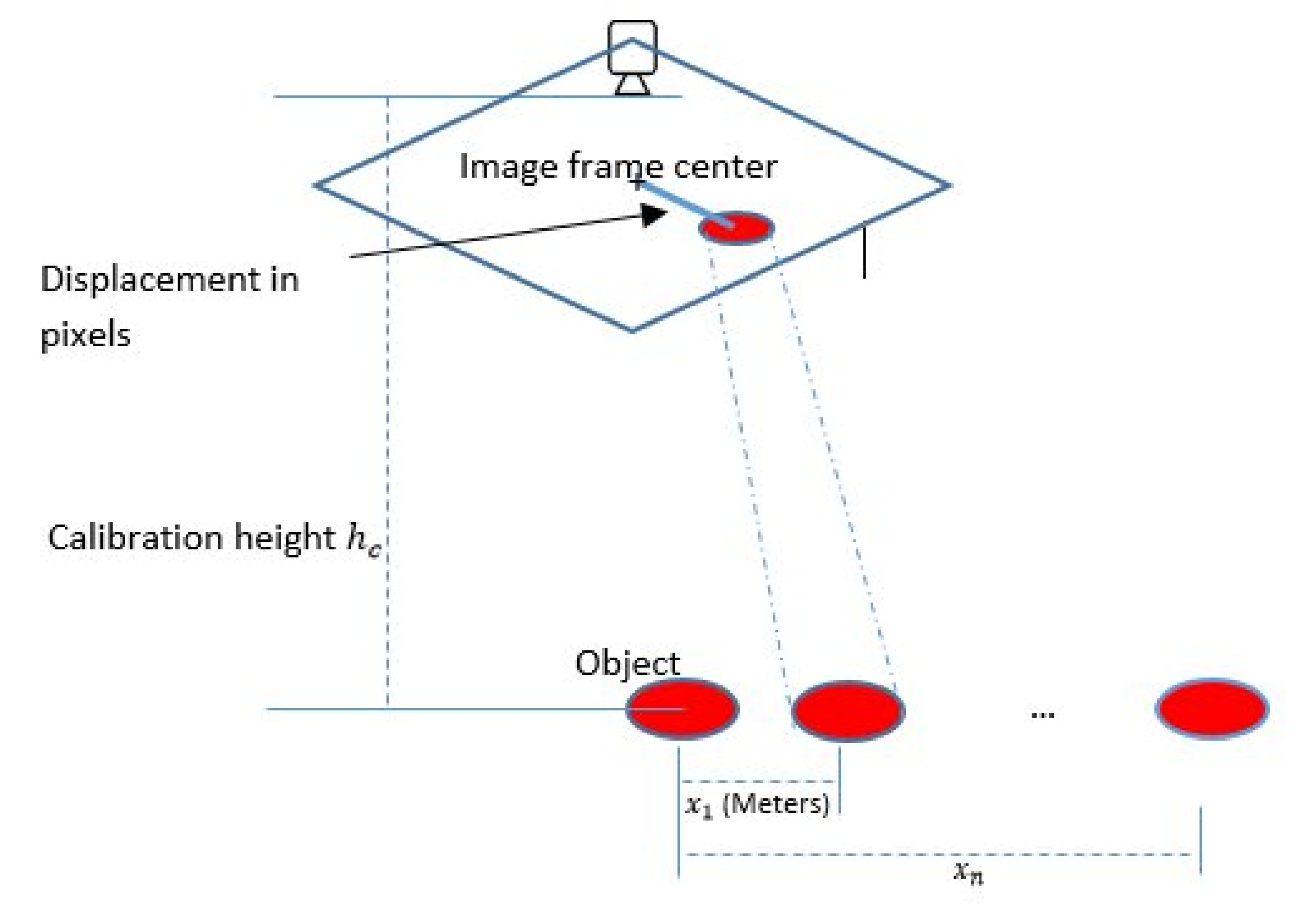}    
\caption{Camera calibration setup: Camera is fixed at a specific calibration height $h_c$. In each trial, the object is placed at a different location and the corresponding physical position from the camera center is recorded. In addition, the pixel displacement of the corresponding object center with respect to the center of the image frame is recorded. Finally, an empirical model is derived.} 
\label{fig:camera_calibration}
\end{center}
\end{figure}

The calibration process (see Fig.~\ref{fig:camera_calibration}) proceeds as follows. A camera sensor is fixed at a known altitude from an object of interest. Then, the object is horizontally displaced with known distances ($x_1, \cdots, x_n$ in Fig.~\ref{fig:camera_calibration}) from the camera center. At each displacement, the reported radial pixel displacement to the object center is recorded against the corresponding actual displacement in meters. Using a specific camera sensor, a table of several measurements is constructed and a fitting function is derived. Following is a quadratic approximation of the relationship between the radial pixel displacement in the image frame and the estimated position of the object, in meters, relative to the UAV coordinate frame.

\begin{equation}
\begin{aligned}
dx_{h_c} &= 0.0018037 x_{\text{pixels}}^2 + 0.3124266 x_{\text{pixels}}\\
dy_{h_c} &= 0.0018037 y_{\text{pixels}}^2 + 0.3124266 y_{\text{pixels}}
\end{aligned}
\end{equation}
where $h_c$ is the calibration altitude, $dx_{h_c}, dy_{h_c}$ are the estimated object distances relative to the UAV in meters at the calibration altitude, and ($x_{\text{pixels}}, y_{\text{pixels}}$) is the detected object center in the horizontal image frame in pixels. In order to adapt to object localization at different altitudes, the estimated distances $dx_{h_c}, dy_{h_c}$ are linearly scaled according to the ratio of the actual altitude to the calibration altitude. That is,

\begin{equation}
\begin{aligned}
dx &= \frac{h_{\text{actual}}}{h_c} dx_{h_c}\\
dy &= \frac{h_{\text{actual}}}{h_c} dy_{h_c}
\end{aligned}
\end{equation}

\begin{figure}
\begin{center}
\includegraphics[width=8.4cm]{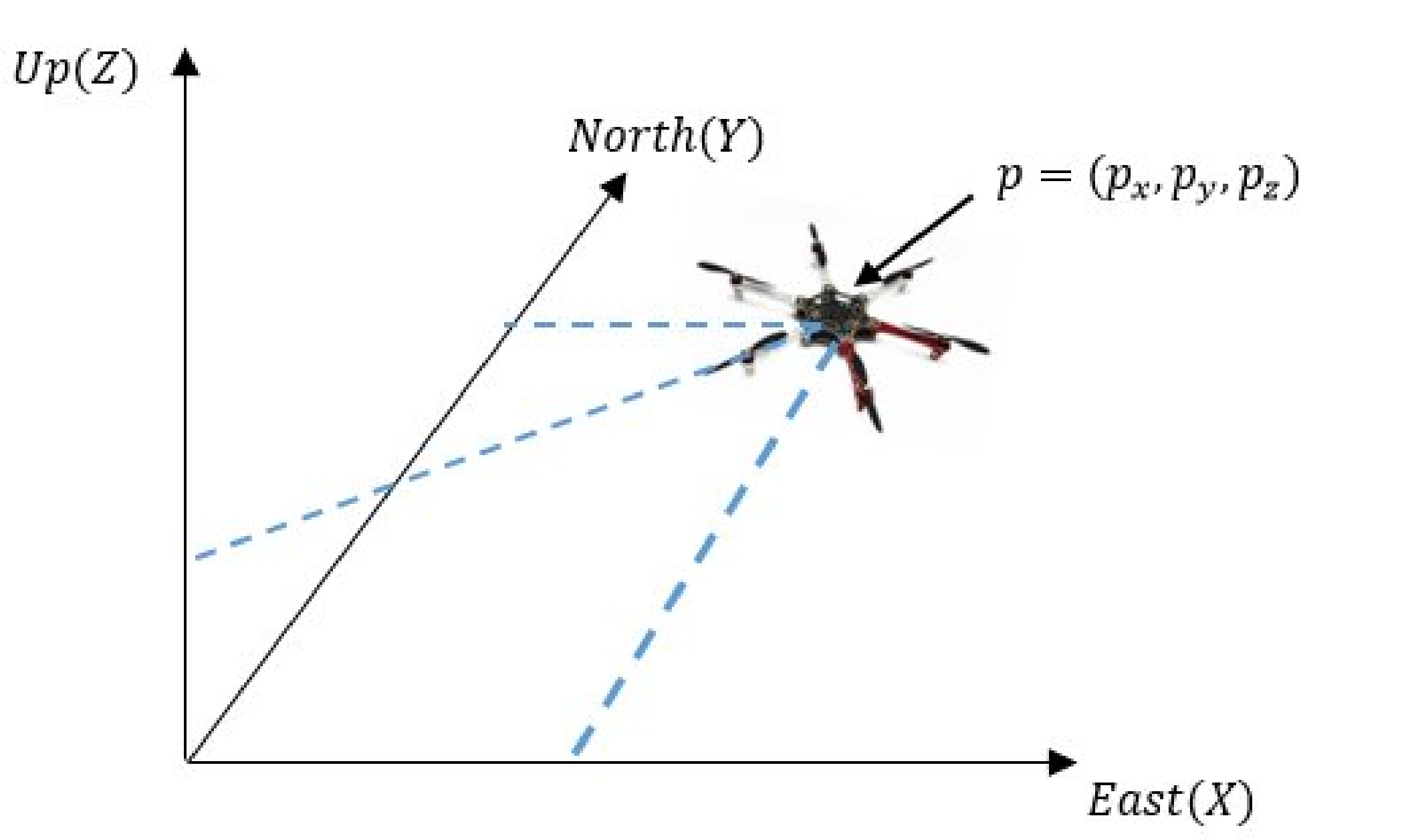}    
\caption{UAV position $p$ defined in a local fixed ENU frame.} 
\label{fig:fixed_enu}
\end{center}
\end{figure}

\subsection{Obtaining Control Set-points}
\begin{figure}
\begin{center}
\includegraphics[width=10cm]{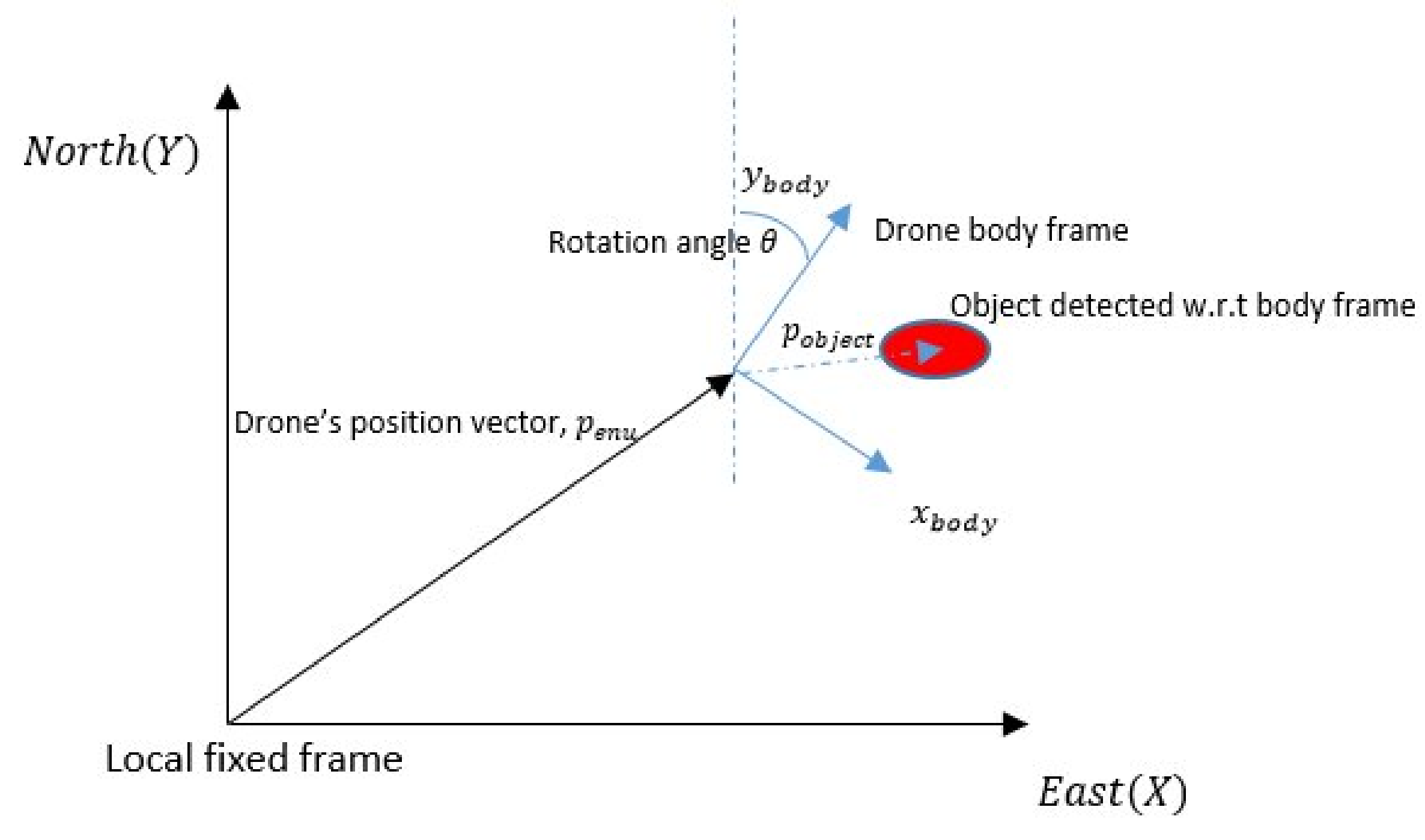}    
\caption{UAV body frame with y-axis pointing in the forward direction. The object is detected with respect to the body frame.} 
\label{fig:body_enu}
\end{center}
\end{figure}

We use position set-points to navigate the UAV towards a detected object center. The UAV position is defined with respect to a fixed local coordinate frame, called $L_F$ of ENU convention i.e., East($X$), North($Y$), Up($Z$). (see Fig.~\ref{fig:fixed_enu}). For the UAV, we define a fixed body frame, i.e., $B_F$. The flight controller takes position set-points $p_{sp}=(x_{sp}, y_{sp}, z_{sp})$ in the local fixed, $L_F$ frame. However, the estimated position of the object is in the $B_F$ frame. Therefore, a transformation of the object position from $B_F$ to $L_F$ is required in order to obtain a valid UAV position set-point. This is achieved by:
\begin{itemize}
	\item A rotational transformation using the body rotation angle $\theta$ (see Fig~\ref{fig:body_enu}), to align $B_F$ with $L_F$.
	\item A translational transformation to finally express object position in $L_F$.
\end{itemize}

Let $p_{\text{enu}} \in R^2$ denote the UAV position in the local fixed ENU frame $L_F$, $r=(r_x, r_y) \in R^2$ the object position with respect to the UAV body frame $B_F$, $r(\theta)=(r_x(\theta), r_y(\theta))$ the object position with respect to body frame after a rotation of $\theta$ around the body frame $z$-axis, and $r_{\text{enu}}$ the object position expressed in the local fixed frame $L_F$ after a translational transformation. Both transformations are done in Eq.~\ref{eq:object_trans} as given below:

\begin{equation}
\label{eq:object_trans}
\begin{aligned}
r_x(\theta) &= r_y sin(\theta) + r_x cos(\theta)\\
r_y(\theta) &= r_y cos(\theta) - r_X sin(\theta)\\
r_{\text{enu}} &= p_{\text{enu}} + r(\theta)\\
\end{aligned}
\end{equation}

Finally, the flight control set-point is simply $r_{\text{enu}}$ which drives the UAV to a new position towards the detected object.

\subsection{Color Thresholding}

Another essential part of solving this problem is a reliable and versatile methodology for detecting colored objects. To this end, we have developed a simple yet effective strategy that also allows for user input for very fast online calibration of the vision algorithm for object detection.

The appearance of the objects outdoors can vary significantly due to environment variations such as time of day, weather conditions, etc. Hence, we design a method that does not require any training data but only requires tuning of a few threshold parameters. In essence, we simply threshold the input image in different color spaces and then merge the results. The thresholds for each color space are determined in a semi-automatic fashion. The user points the camera at a colored object and provides a tolerance threshold to determine the sensitivity of the determined thresholds. The thresholds for each color space are then determined automatically. This procedure is repeated for each color and the determined thresholds are saved to a local configuration file and synchronized with the ROS server. 

\begin{figure}
\begin{center}
\includegraphics[width=12cm]{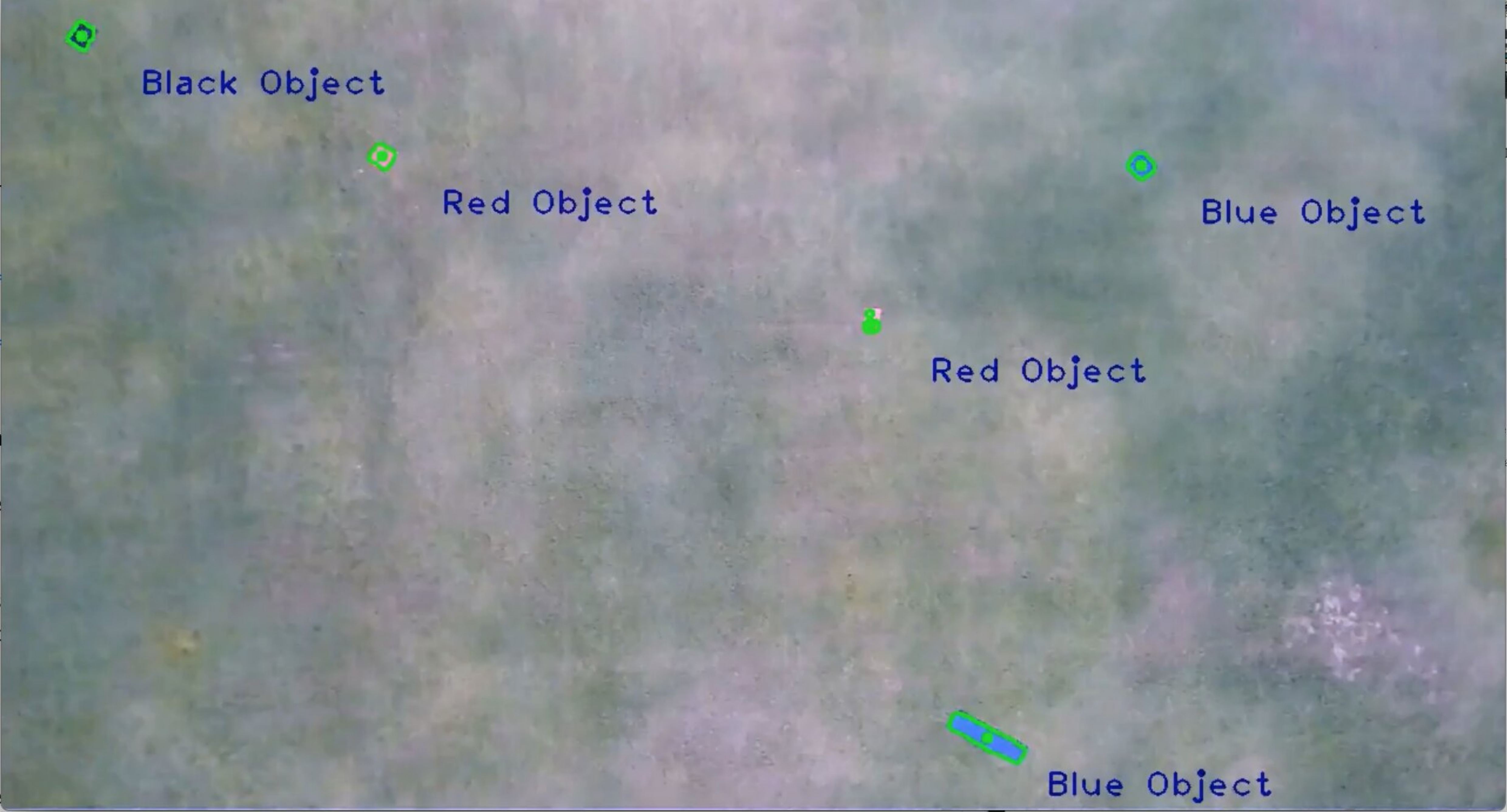}    
\caption{Object detection using color thresholding in various color spaces.} 
\label{fig:object_detection}
\end{center}
\end{figure}

Extensive experiments show that the LAB color space provides the best separation of the colors used in this challenge (blue, green, red, yellow, orange). In addition, we use the HLS color space which provides some invariance to illumination, and lastly the RBG color space in which the images are captured. We combine the thresholded images for each color space into a single RGB image where each channel now corresponds to a thresholded image. We then convert this RGB image to a gray scale image. The color channels are weighted when converting to gray scale, effectively providing automatic weights for the different color spaces (HSL - 0.2989, LAB - 0.5870, RGB - 0.1140). The merged result now contains a thresholded image for a specific color with very little noise due to this smart combination of different color spaces. We then find the contours on this thresholded image and fit the appropriate shapes (e.g., circle in our case). This methodology is very efficient and achieves close to real-time performance on an embedded platform such as the Odroid XU-4. It can be tuned very quickly and not only that it works well for detecting and tracking the colored objects, but also for localizing the rectangular drop zone precisely (see Fig.~\ref{fig:object_detection}).

\section{Aerial Grasping and Transport}\label{sec:grasping}

In this section, we present a simple light-weight gripping mechanism for ferrous objects with feedback on the picking state. The mechanism is based on our novel design for passive aerial grasping and transport of ferrous objects \citep{fiaz2019impulsive}. We also describe its actuation routine and a reliable picking strategy for grasping objects outdoors even in presence of high wind disturbance.

\subsection{The Grasping and Release Mechanism}

\begin{figure}
\begin{center}
\includegraphics[width=12cm]{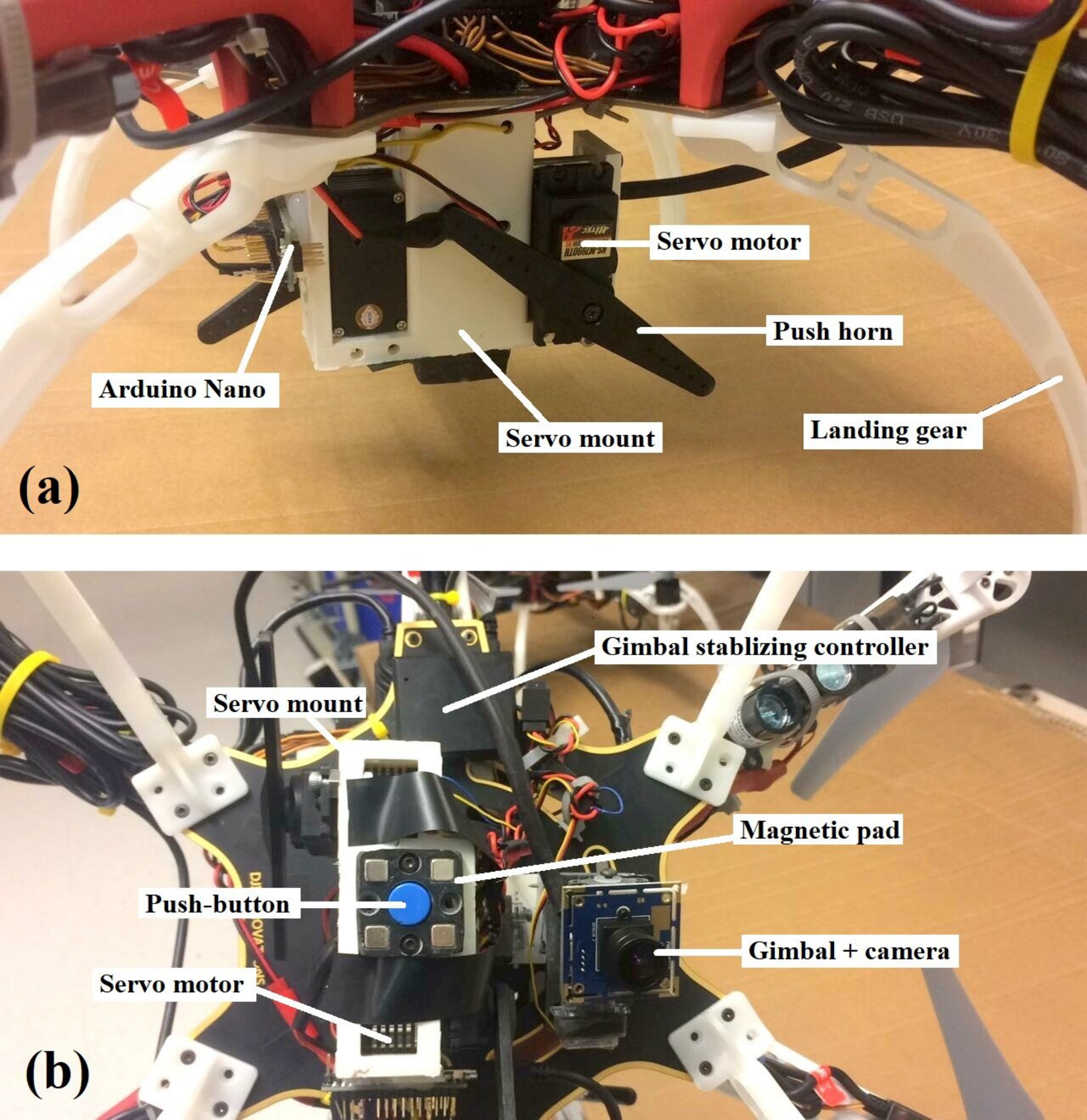}    
\caption{Gripper design: (a) side view, and (b) bottom view. The 3D printed gripper enclosure holds together two servo motors, four permanent magnets, push-button for feedback, gimbaled camera with its holder, and Arduino Nano for actuation control and ROS interface.}
\label{fig:gripper}
\end{center}
\end{figure}

Payload is an important consideration while designing a gripper for drones. We would like to keep the grip as strong as possible while keeping the mechanism weight to a minimum. Thus, for ferrous grasping application, we investigated various options including electromagnets, electro-permanent magnets (EPMs), and permanent magnets. Low power consumption compared to electromagnets, high payload capability, and convenient commercial availability of the EPMs, apparently makes them a default choice. However, EPMs are shown to have problems with flushing on to the surface of the objects on touchdown, since they require a few seconds to activate in order to grasp a ferrous payload with full strength, and need perfect alignment with the payload surface \citep{fiaz2018intelligent}. Therefore, instead we designed our own magnetic gripper with permanent magnets and a novel impulsive, servo-actuated release mechanism, which outperforms EPM based designs. Figure~\ref{fig:gripper} shows the complete gripper assembly mounted to the hexarotor frame. 

All the assembly parts have been designed and printed via the Objet30 Prime 3D printer at the RISC Lab. The whole gripper when assembled, weighs around 250 g. The servo mount holds everything in place. The square magnetic pad at the heart of the mechanism is the key to spontaneous grasping. It employs four 6.33 mm cubes of N42 Neodymium magnets. These magnets are collectively capable of providing a net lift of around 0.76kg. For our experiments, the test objects we used weigh 500 g at maximum. Thus, one pad does the job for us. The pad also contains in its center, a push-button, that is pressed and released every time the gripper picks up and drops an object respectively. As is described later in this section, this little feature is vital for ensuring flawless autonomous flow of the finite state machine (FSM) during the grasping operation. 

\begin{figure}[h]
\begin{center}
\includegraphics[width=8.4cm]{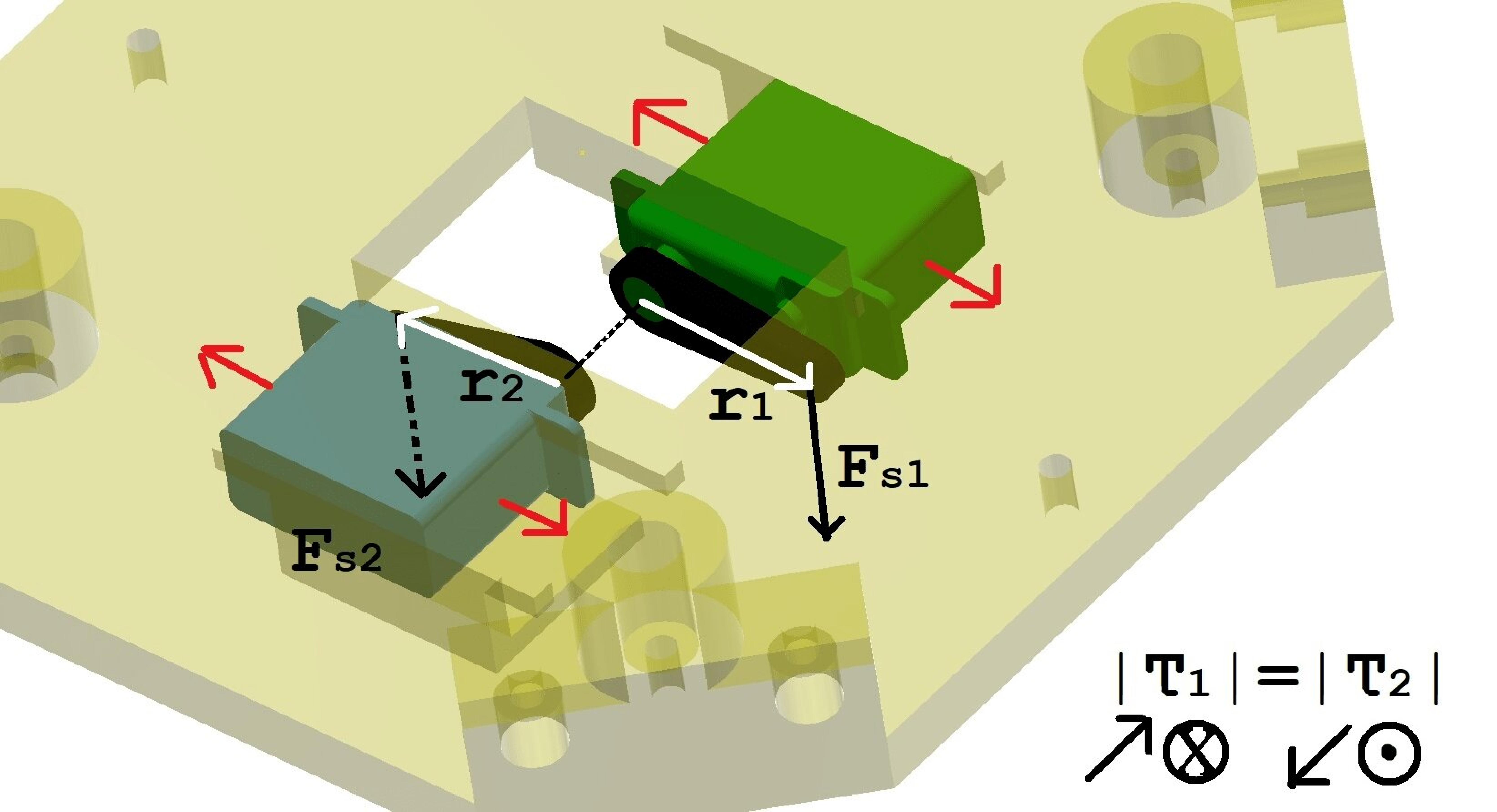}    
\caption{Dual impulsive release mechanism with counter-torque. Two equal and opposing torques of the servo actuators double the release force on the ferrous payload attached to the magnets while preventing any torsional effect in the gripper assembly \citep{fiaz2018intelligent}.}
\label{fig:release}
\end{center}
\end{figure}

The release mechanism as shown in Fig.~\ref{fig:gripper} consists of two high speed servo actuators, which when activated push the object off the magnetic pad using their respective horns. The two servos are mounted at right angles to each other ensuring a counter-torque (see \cite{fiaz2018intelligent}), when activated at the same time. This concept of dual impulsive release (see Fig.~\ref{fig:release}) is quite efficient in terms of design simplicity as well as power consumption, since the only time the gripper consumes power is in the drop phase. The average power consumption over a complete pick and drop cycle of the gripper operation is only 3.48 W. 

An Arduino Nano serves as a dedicated ROS node for controlling the gripper actuation. It reads the push-button feedback from the magnetic pad and publishes the pick/drop status to Odroid (i.e., the companion computer) in real time. It is subscribed to pick/drop commands from the Odroid as well, in response to which it either activates or deactivates the release (servo) mechanism.

\subsection{Camera Stabilization}

\begin{figure}[h]
\begin{center}
\includegraphics[width=7cm]{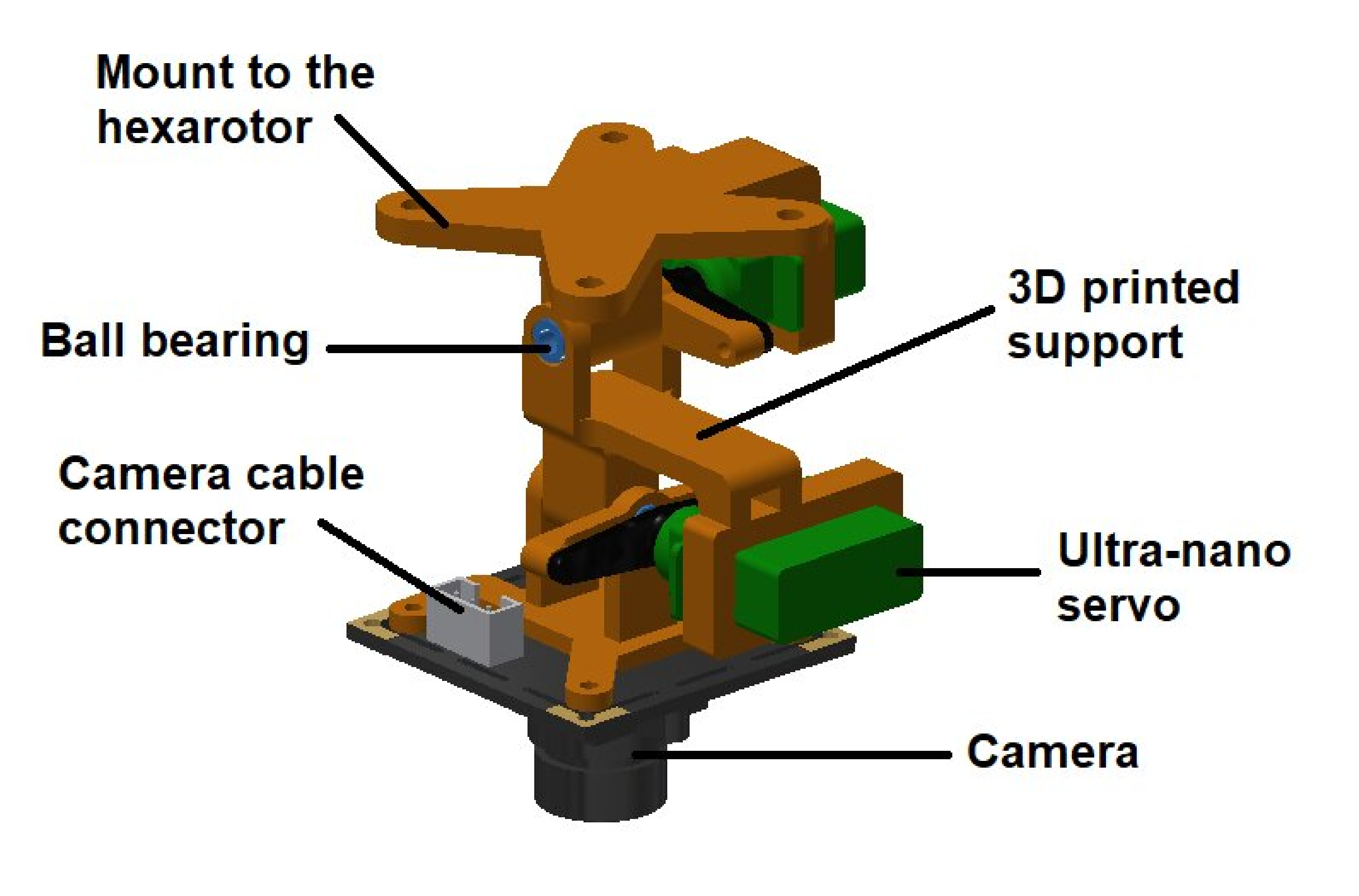}    
\caption{CAD animation of the 3D printed ultra-nano stabilization gimbal for the camera.}
\label{fig:gimbal}
\end{center}
\end{figure}

In addition to the grasping and release mechanism, the gripper assembly also has a built-in ultra-nano servo gimbal for the camera module (see Fig.~\ref{fig:gimbal}). This customized 3D printed gimbal uses two Hitech ultra-nano servos to stabilize the roll and pitch of the camera as the UAV flies and carries out various maneuvers. This keeps the camera faced down, aligned with ground all the time, which makes the object detection and localization convenient.

\subsection{Actuation Routine}
Each of the three UAVs in our testbed is equipped with an identical gripper assembly. The actuation and grasping routine for any UAV proceeds as follows: The magnets being permanent are activated by default. In the picking state, the servos are deactivated i.e., the horns rest above the magnetic pad. Thus as a UAV detects, descends and picks up an object, the feedback signal from the push-button switches from 0 to 1. A 0 means an object is not picked, while a 1 means that an object has been picked up successfully. Thus a 1 message serves as a pick up confirmation for the FSM. Now, when a UAV reaches the drop zone, the Arduino (ROS node) receives a drop signal from the Odroid (FSM), and hence it activates the release mechanism. As the object is dropped, the push-button feedback switches from 1 to 0. Similar to the picking routine, a 0 message serves as the drop confirmation for the FSM. Once it gets the confirmation, it proceeds to the next state and also sends a pick up signal to the Arduino which deactivates the release mechanism again, and the process continues.

\subsection{Picking Strategy}

One of the main contributions of this work is our simple yet reliable picking strategy, which is the way the drone will approach the object to be picked. The proposed picking strategy relies on accurate tracking of the estimated object position, based on vision and the MAV altitude from ground. As stated earlier, a LiDAR sensor is used for accurate altitude estimation. For vision based object localization, however, objects can not be always detected in all image frames due to environmental conditions and disturbances. For this reason, we adopt a confidence based approach to descend towards an object only if there is high confidence that it is detected and centered within a certain region.

\begin{figure}
\begin{center}
\includegraphics[width=7cm]{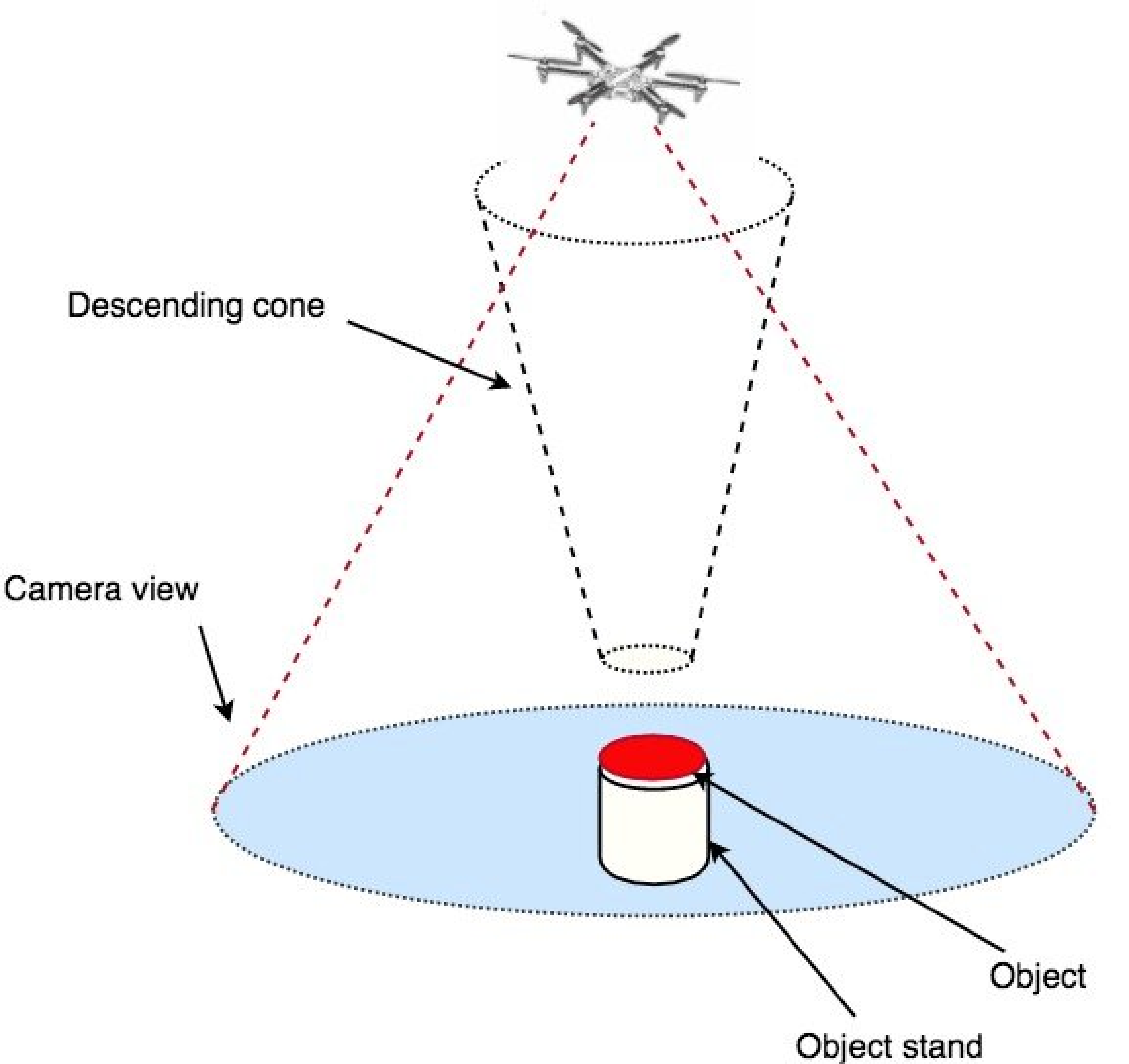}   
\caption{Descending cone for a UAV during the picking state.} 
\label{fig:des_cone}
\end{center}
\end{figure}

\begin{figure}
\begin{center}
\includegraphics[width=12.0cm]{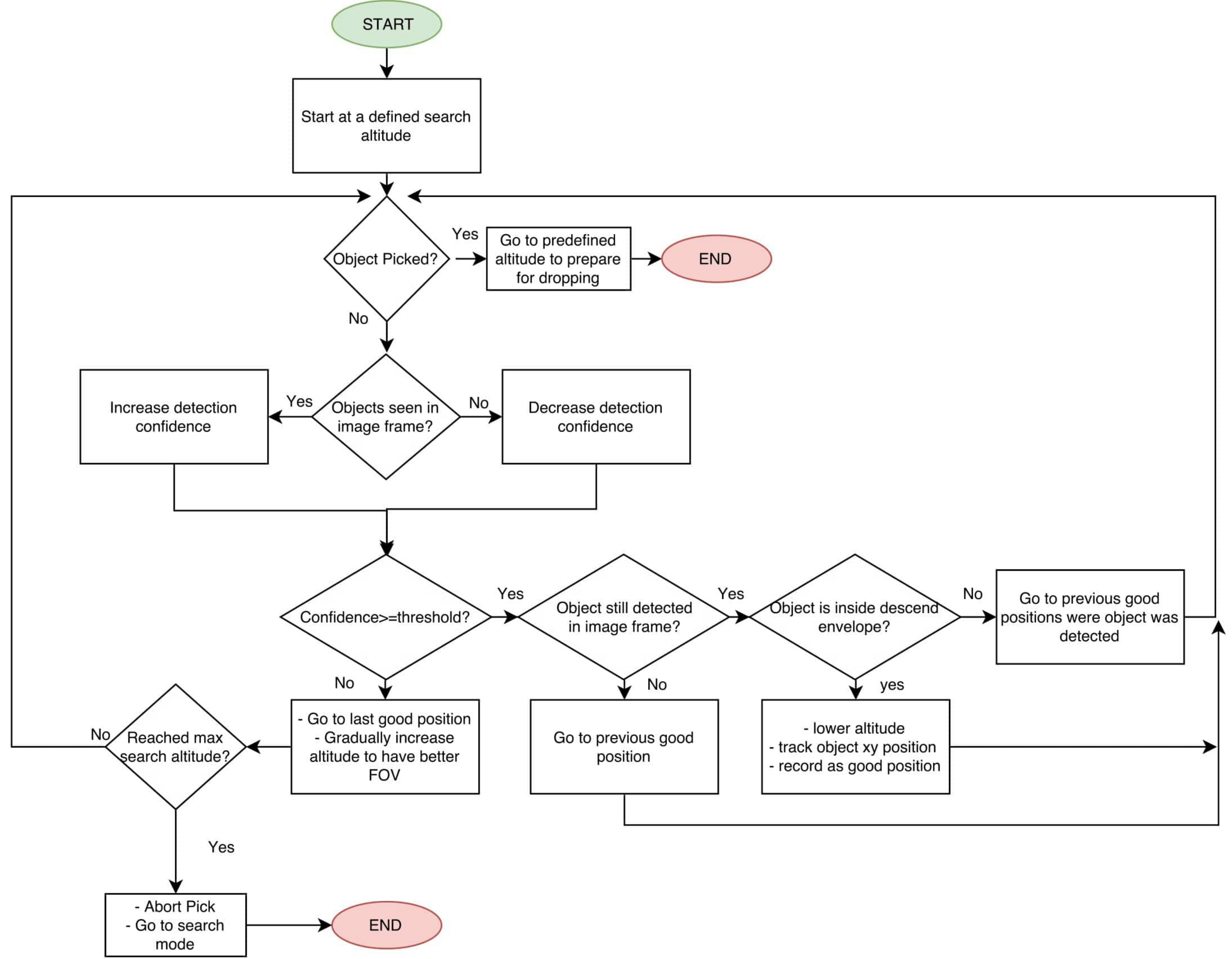}    
\caption{Picking state machine for a UAV.} 
\label{fig:pick_sm}
\end{center}
\end{figure}

Our picking strategy works as follows. First, once an object is detected, the UAV is commanded to do lateral tracking of the object based on the estimated object position using vision. A confidence parameter is updated based on the frequency of detection in image frames. Next, if the confidence is higher than a predefined threshold and the UAV is within a certain vicinity around the object, the altitude is decreased gradually (see the code for details). The vicinity threshold at which the UAV is considered safe to descend, is defined by a cone with decreasing radius as shown in Fig.~\ref{fig:des_cone}. This allows the UAV to descend more quickly when at a high altitude while being conservative at low altitude, for accurate positioning onto the object center. If the confidence is low, the UAV falls back to a good altitude where it last saw the object. This approach is encoded in yet another finite state machine shown in Fig.~\ref{fig:pick_sm}. This approach proved to provide accurate and smooth centering over the detected objects in the field experiments, which we discuss in Section~\ref{sec:experiments}.

\section{Communication}\label{sec:comms}

In our multi-UAV testbed, we use WiFi enabled communication between the UAVs using a dedicated 2.4 GHz outdoor network. Thanks to the partitioning approach, we require communication between the UAVs only during the dropping phase. Even then, UAVs only need to share simple data such as their current state (e.g. takeoff, picking, dropping, .. etc.) and position with one another to avoid collisions. In this section, we describe a simple software application that uses a custom MAVLink message for intercommunication between the three UAVs. The MAVLink protocol and its simple message customization provide a reliable encoding/decoding mechanism as well as make the handling of communicated messages rather trivial.

\begin{figure}
\begin{center}
\includegraphics[width=10cm]{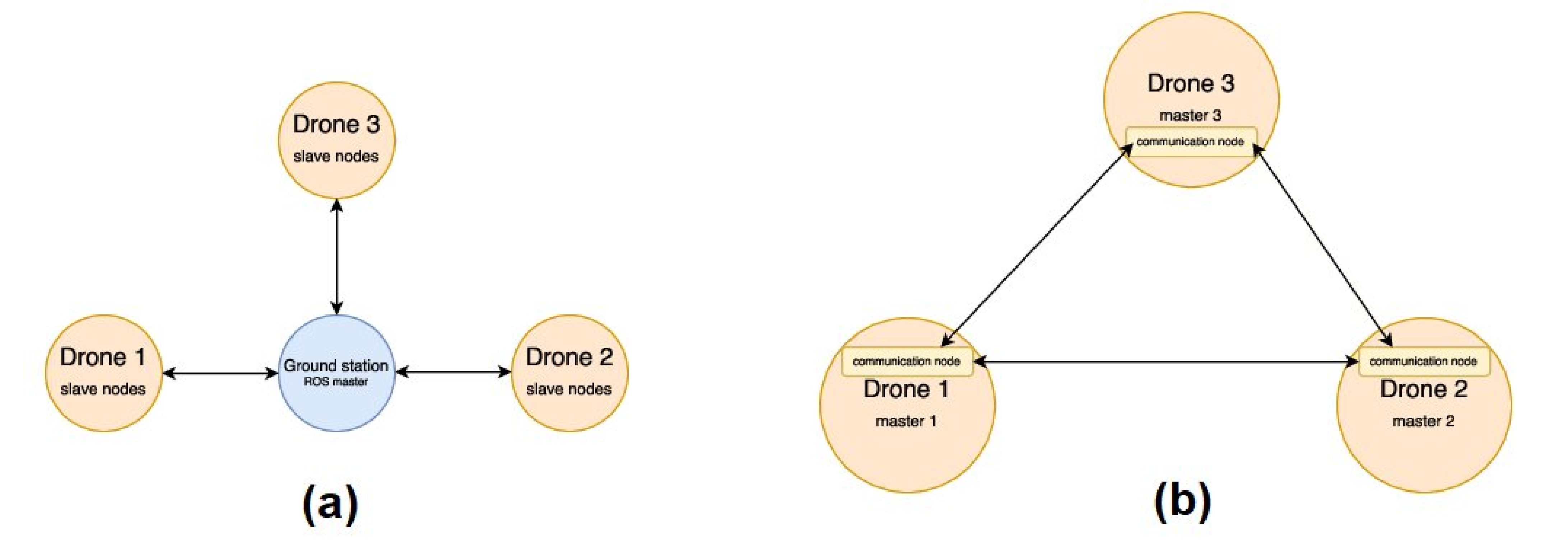}   
\caption{Inter-UAV communication architecture: Figure (a) shows the standard ROS communication way with a centralized master node, which is prone to single point of failure issue. Figure (b) shows our customized communication method with distributed master nodes, to mitigate this issue.} 
\label{fig:ros_arch}
\end{center}
\end{figure}

As is emphasized earlier, due to a limited space of the drop zone, it is necessary to guarantee collision free drop of the objects, in case more than one UAVs are in the drop state at the same time. Although a vision based approach may be feasible for a UAV to identify a partner drone \citep{6871124}, \citep{7759252} in the drop zone, it will require extra computation and tuning to reach a satisfactory level of robustness and reliability. To simplify this task, we use communication enabled consensus to share simple information, e.g., current position and current mission state, between the three UAVs. The role of this communication here is to provide the UAVs with the needed information to do coordinated and collision-free drop. In our system, we use customized communication programs (ROS nodes) in order to allow the UAVs to have their independent ROS master node, which greatly reduces the chances of failure of the overall system. Figure~\ref{fig:ros_arch} elaborates this idea in pictorial form.

Our software architecture is running on top of ROS which in principle, allows for setting up a distributed system. However, for that to work, only one machine has to be defined as the master node which runs the ROS core communication interface, which is responsible for connecting other nodes together, either on the same machine or others (see Fig.~\ref{fig:ros_arch}(a)). If the communication structure is reliable, e.g., a reliable transmission through the physical WiFi setup, the standard ROS communication architecture will work perfectly fine, and all nodes can easily share their information through the master node. However, if one node fails to communicate to the master node at some point in time, the node execution is affected and can be interrupted, and eventually can lead to a node crash. In fact, we faced such problems when only one master node was used in our experiments; i.e., the ground control station (GCS) computer was the master node, and all three drones were connected to it as slave nodes. A major problem would arise, whenever a drone lost connection to the master node on the GCS, and the node execution would be interrupted, which in turn would lead to mission interruptions.

In order to solve this issue, we resort to a distributed master node architecture (see Fig.~\ref{fig:ros_arch}(b)). This is achieved by letting each drone run its own master ROS core node locally, in order to avoid the dependency on a remote master and the occasional disconnections resulting from it. However, using this method, other UAVs information (called topics in ROS terminology) are not available anymore, and a special communication pipeline is needed. Therefore, we customized a simple ROS node on each drone to handle the communication of the required information, i.e., positions and mission states using the UDP protocol. We chose the UDP protocol as it does not involve handshaking mechanisms, which reduces latency and increases the data throughput.

\begin{figure}[H]
\begin{center}
\includegraphics[width=10cm]{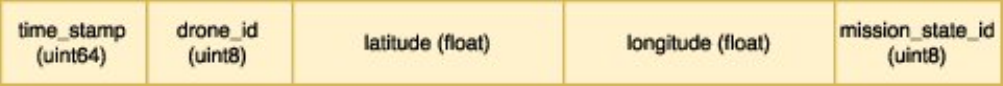}    
\caption{Content of custom MAVLink message used in inter-UAV communication.} 
\label{fig:mavlink_msg}
\end{center}
\end{figure}

Each communication node performs two tasks. First, it subscribes to its UAV position and mission state, encodes them in a customized MAVLink message, and sends it to the other UAVs. Secondly, it listens to messages form other UAVs, decodes them using the same definition of the custom MAVLink message, and publishes them locally as ROS topics to be used by other local nodes. The MAVLink protocol is used because it provides a simple way of defining custom messages as well as simple encoding and decoding functionalities. It also includes a checksum in the low-level message construction that helps to recover correct information. The custom MAVLink message packet has a payload of $14$ bytes which includes a UAV-ID, latitude and longitude, and mission state information (e.g., takeoff, picking, dropping etc.). Figure~\ref{fig:mavlink_msg} shows the contents of this message.

\section{Experiments and Results}\label{sec:experiments}
Now, we describe the simulation environment that we used for verifying the mission execution and results from outdoor experiments on the real system.

\subsection{Simulations}

Before experimenting with the physical testbed outdoors, we verified our approach inside a simulation environment. We used V-REP simulator for testing the successful completion of the mission using an identical three UAV system. The only key difference between the simulation and reality is that we used quadrotor UAVs in the V-REP environment, which does not affect the mission results significantly. These simulations helped us a lot in tuning color thresolding parameters for detecting objects of various colors and in determining efficient scanning schemes for the UAV partitions. It also enabled us to verify the correct execution of the FSM for the mission. 

\begin{figure}[]
\begin{center}
\includegraphics[width=12cm]{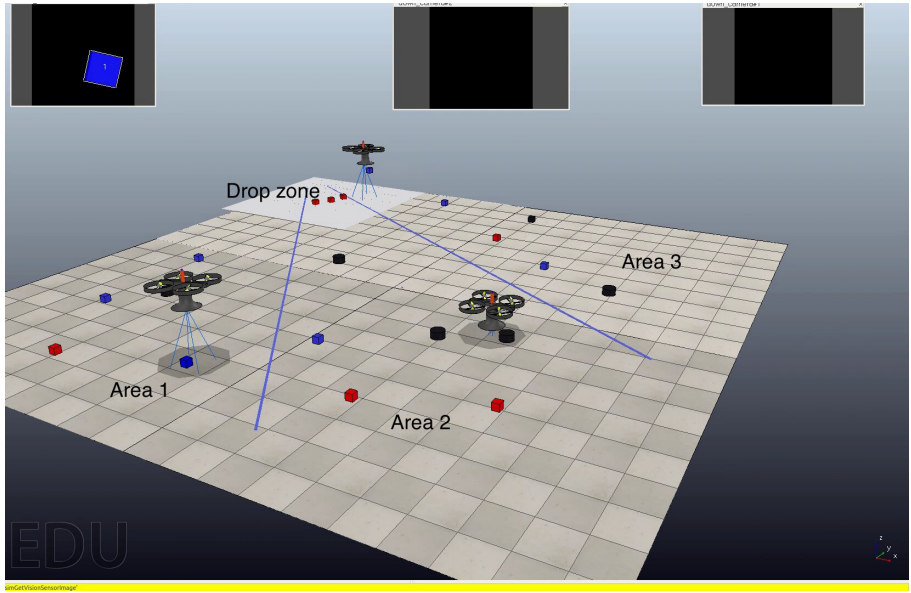}    
\caption{An example of the system setup while being simulated in the V-REP robotic simulator, where three identical UAVs are considered. The search area is divided into three partitions as before, and colored objects are placed randomly in the search space. The downward facing camera view for each drone is shown in separate windows on the top.} 
\label{fig:simulation}
\end{center}
\end{figure}

After several successful simulation runs in V-REP, and after fine tuning of the FSM and the vision parameters, we were ready to do outdoor experiments with the real testbed. A snapshot of the V-REP multi-UAV simulation is shown in Fig.~\ref{fig:simulation}. A link to the video of a successful simulation run is also available in Section~\ref{sec:links}.

\subsection{Outdoor Experiments}

In the following, we show results from a single UAV test and a complete experimental run for the search and rescue mission, which demonstrates autonomous exploration, grasping and coordinated transport. A video of the experiment is available in Section~\ref{sec:links} as well.

\begin{figure}
\begin{center}
\includegraphics[width=12cm]{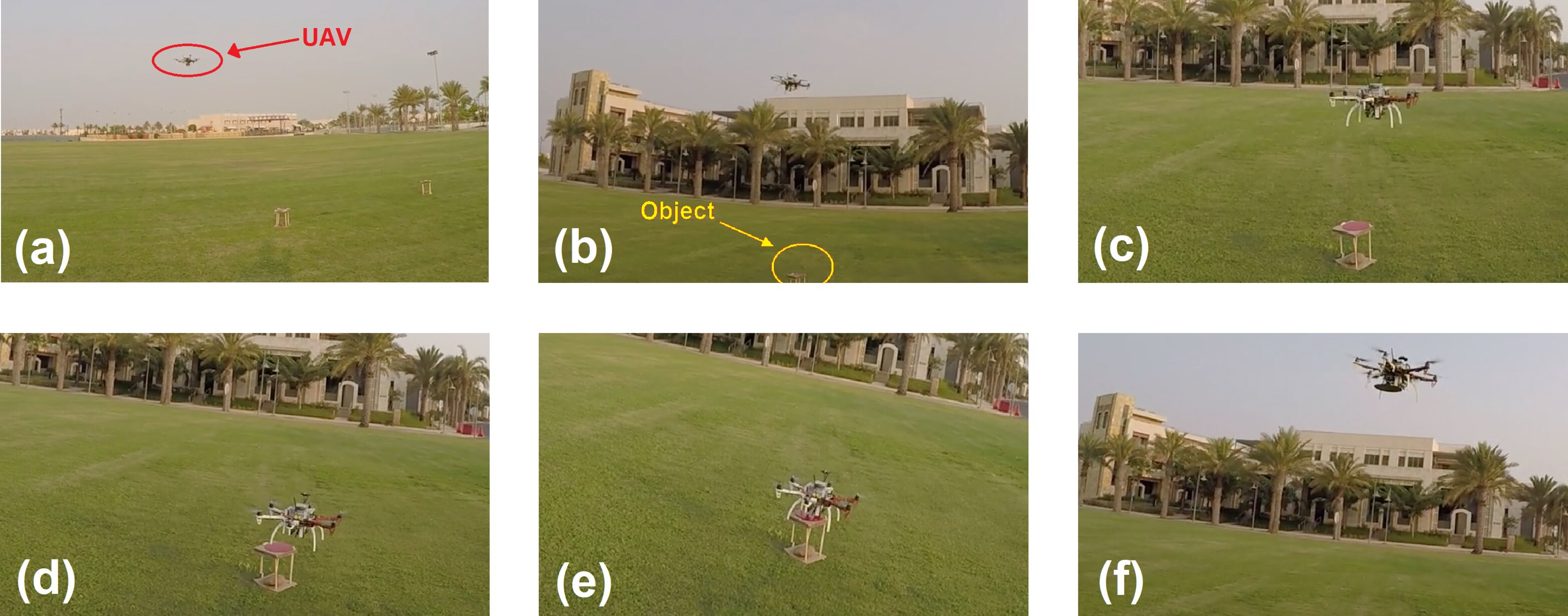}    
\caption{Single UAV testing; (a) a snapshot of the drone in the search phase, (b) a snapshot of the drone after an object is found and selected, (c) a snapshot of the drone while descending over the object, (d)  a snapshot of the drone while aligning over the object to prepare for picking, (e) a snapshot of the drone while picking the object, and (f) a snapshot of the drone after picking the object, going to the drop zone.} 
\label{fig:single_drone_test}
\end{center}
\end{figure}

Each drone was tested individually in order to verify correct execution of each operational task during the mission. This included area exploration, object detection, object picking test, dropping test, and eventually the overall autonomous mission which is managed by the FSM. Figure~\ref{fig:single_drone_test} shows snapshots of testing the individual tasks during an autonomous mission for a single UAV.

\begin{figure}[]
\begin{center}%
\includegraphics[width=12cm]{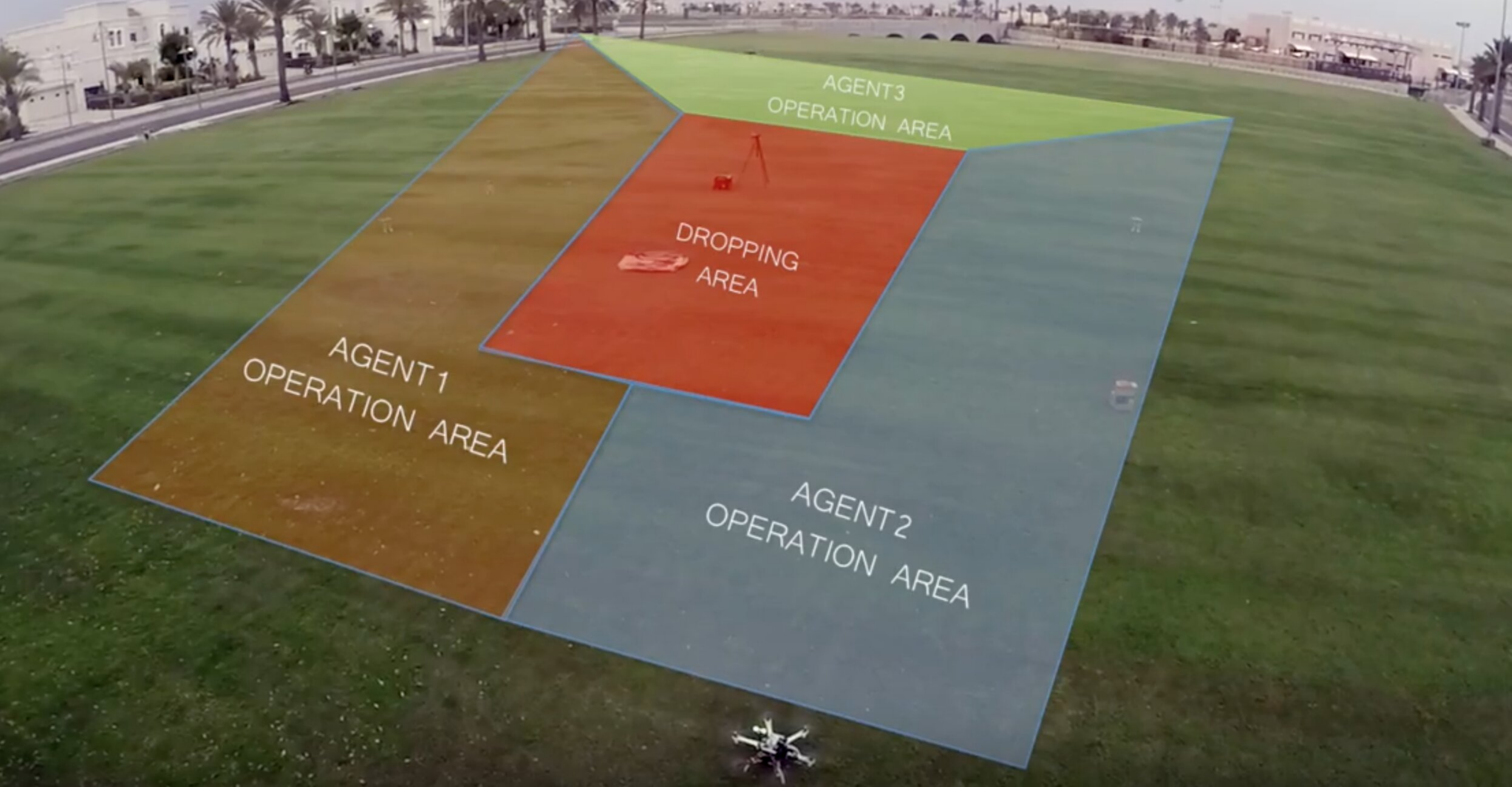}
\caption{A screenshot overlay of field partitioning in the outdoor experiments.}
\label{fig:partitioning}
\end{center}
\end{figure}%

\begin{figure}[]
\begin{center}%
\includegraphics[width=12cm]{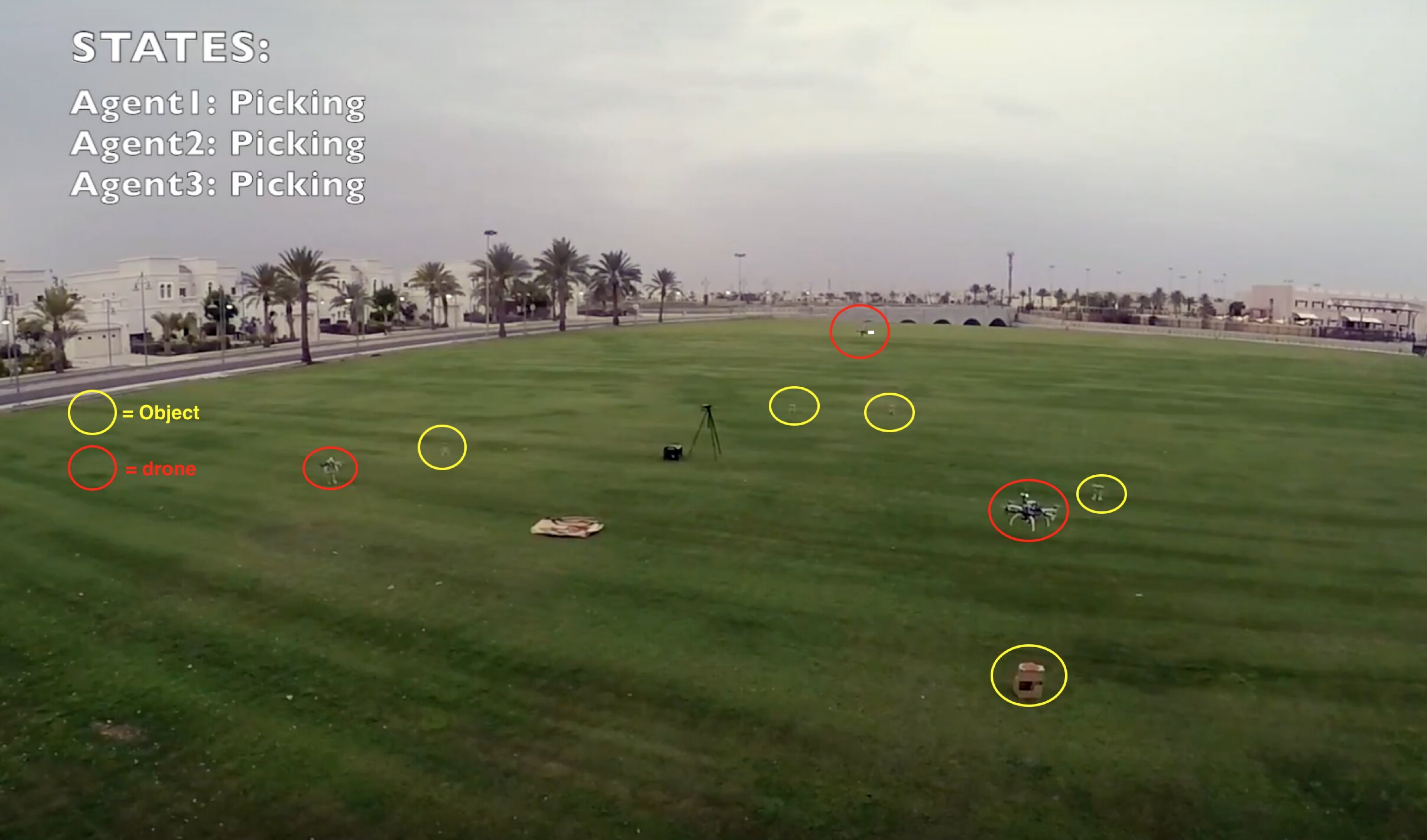}
\caption{Snapshot from the outdoor experiments.} 
\label{fig:grand_test}
\end{center}
\end{figure}

A multi-UAV experiment with a full mission i.e., the RISCuer was then executed, where the field was divided into three partitions as described before. The corner points of each partition were provided for the corresponding UAV only (see Fig.~\ref{fig:partitioning}). In each operation area, two colored ferrous discs with $10$ cm radius were placed on wooden stands of $30$ cm height, at random locations. Then, the UAVs were given a start signal and they executed the complete mission autonomously afterwards. A Linux computer (i.e., the GCS) was used to monitor the mission execution and UAV states remotely. Several runs were performed to confirm the reliable operation of the testbed and and the successful execution of the mission. All runs were successful with an average completion time of about $3$ minutes. Figure~\ref{fig:grand_test} shows a screenshot from these outdoor experiments.

\subsection{Discussion}

\begin{figure}[]
\begin{center}
\includegraphics[width=10cm]{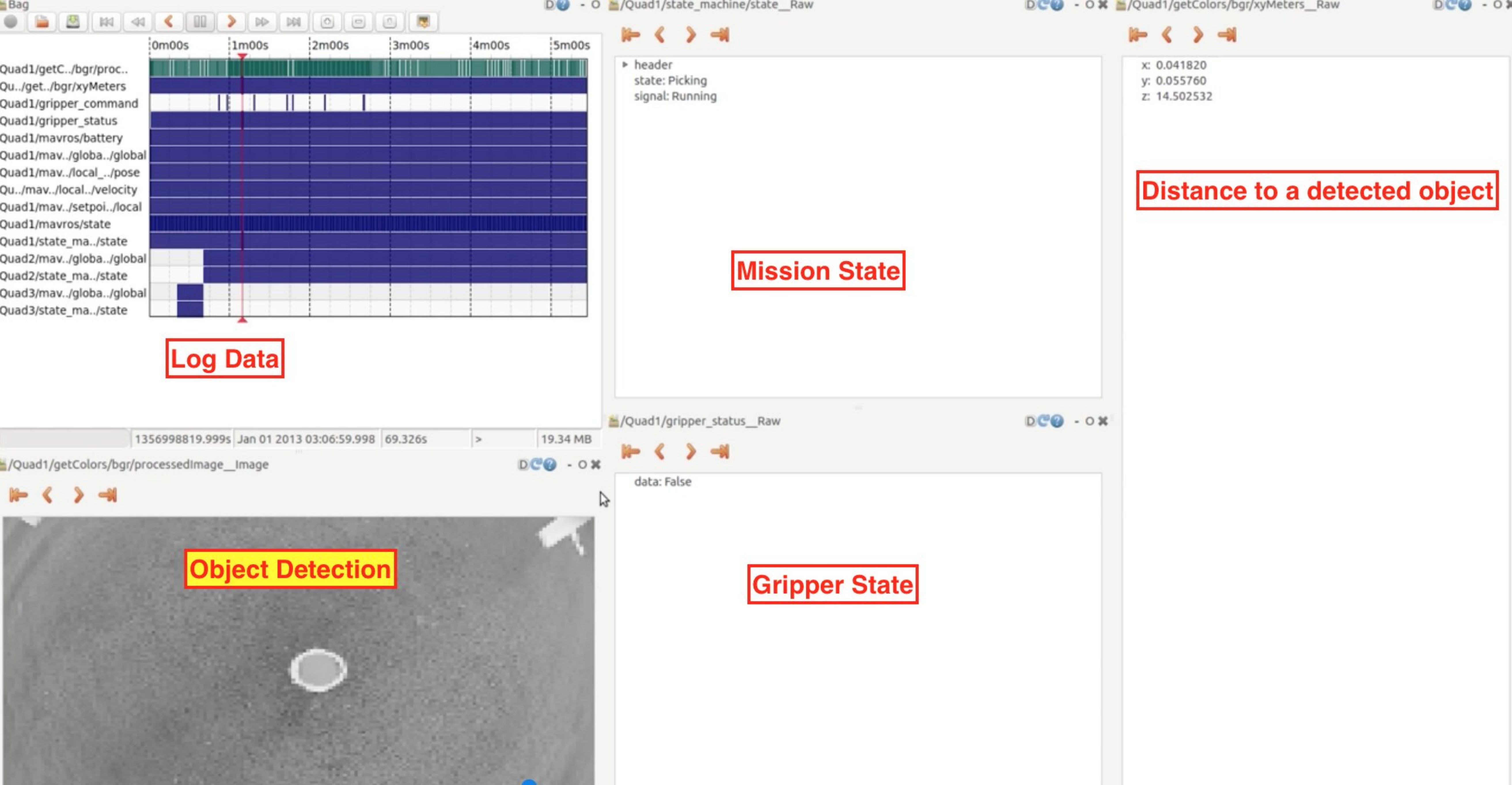}
\caption{Log replay of a drone for a complete mission.}
\label{fig:rosbag}
\end{center}
\end{figure}

In our experiments, we used \textit{rosbags} (data logging system in ROS) for data logging as it provides convenient tools for data visualization and time-stamped mission replays. Figure~\ref{fig:rosbag} shows a snapshot of a log replay of one of the three drones during a complete mission. The logged data includes time-stamped processed gray-scale image where an object is encircled if it is detected, state of the mission, error distance to current detected object, and the gripping status.

\begin{figure}[]
\begin{center}
\includegraphics[width=11.5cm]{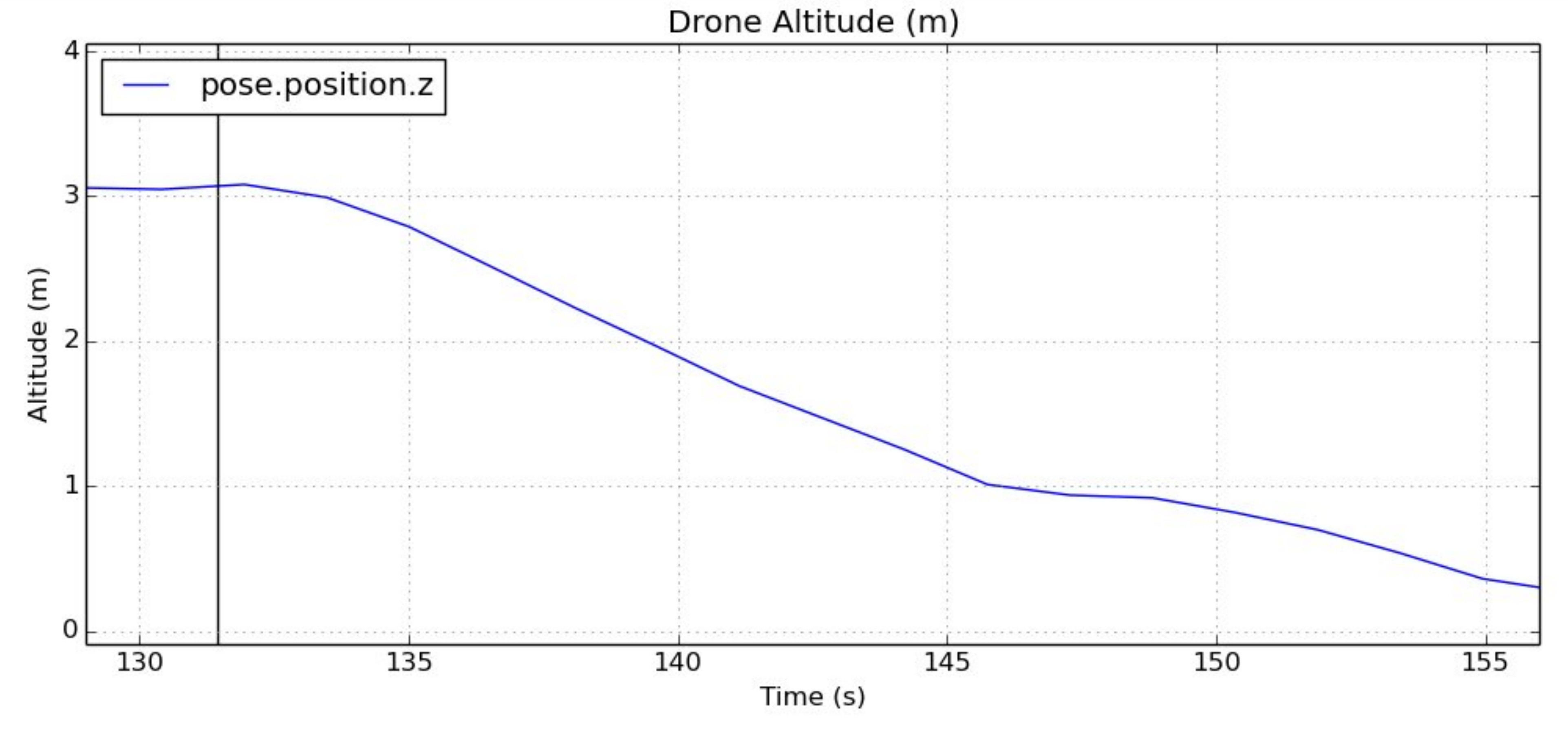}    
\caption{Smooth altitude trajectory of the UAV during the picking state.} 
\label{fig:altitude}
\end{center}
\end{figure}

\begin{figure}
\begin{center}
\includegraphics[width=12cm]{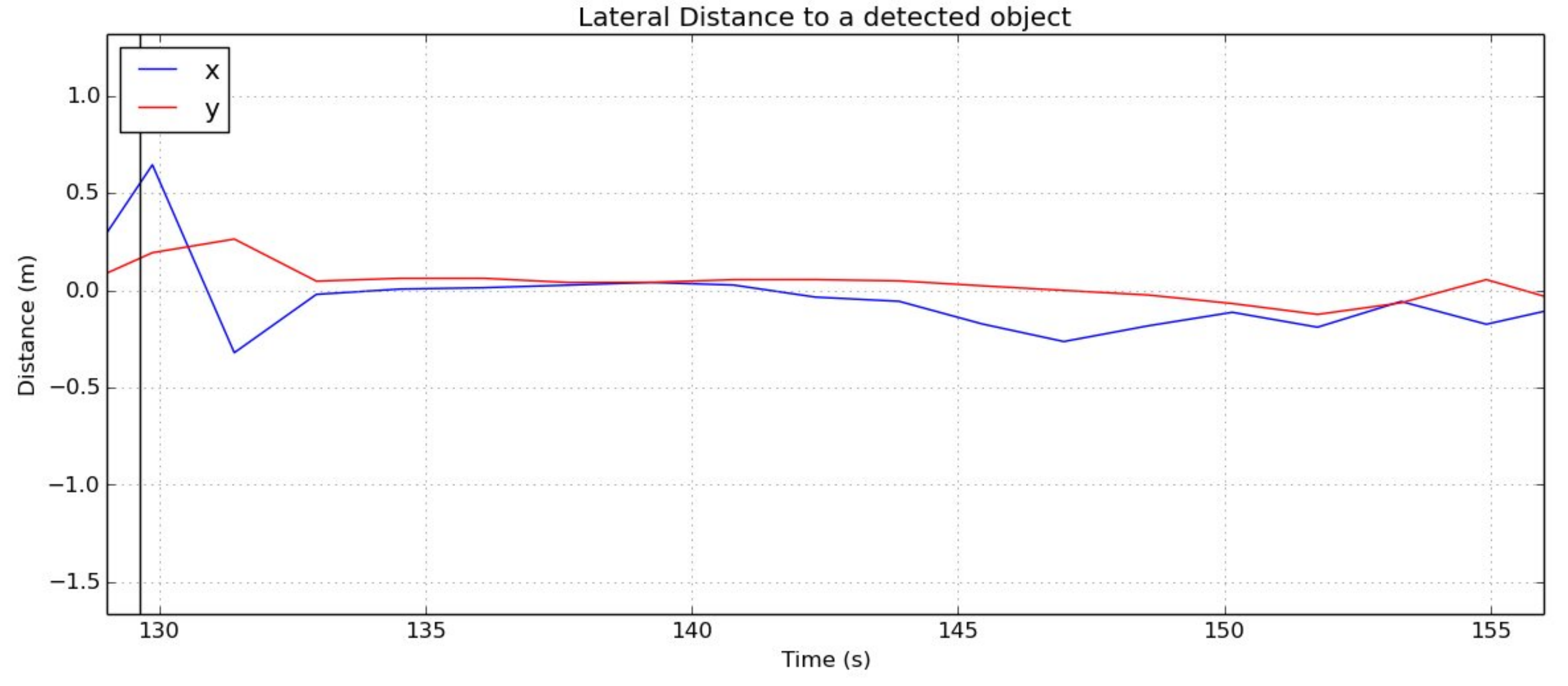}    
\caption{Distance error between the UAV position and the detected object during the picking state.} 
\label{fig:dist2object}
\end{center}
\end{figure}

During several autonomous missions, a main factor of success is the accurate object centering with respect to the UAV gripper, which is a result of an accurate object localization using vision. In Fig.~\ref{fig:altitude}, a smooth descent can be seen as the object is being centered with respect to the gripper center, while Fig.~\ref{fig:dist2object} shows the distance error between the UAV and the detected object during the
picking state. These plots validate the effectiveness of our approach over other recent works, such as \citep{lee2019mission} and \citep{beul2019team}.

The experiments also showed the effectiveness of our proposed grasping mechanism over the EPM based solution. In particular, we performed a comparative study with EPMs in terms of power consumption as well as the payload handling capabilities. Based on this study, the success rates for autonomous pick ups were observed to be 53$\%$ for EPMs and 97$\%$ for our passive design respectively. In addition, the study also showed our mechanism to be more power efficient (see \cite{fiaz2018intelligent} for details). This further strengthens the claim that our system is more reliable than several recent works that use EPMs as their solution for autonomous aerial transport.

We would also like to highlight one of the main challenges that we faced during the course of this work. For the sake of simplicity of the system, we used blob detection methods on low-computation modules for vision based object detection. Such methods are usually tuned for particular colors at specific environmental conditions e.g., light intensity. Therefore, it is challenging to use the same parameters to detect the same colors in different lighting conditions, which we faced during outdoor field tests. More complex methods can be used, but at the expense of more computation power. One possible solution is to use adaptive vision parameters (i.e., color thresholds) according to a pre-trained model which accounts for environmental changes such as light intensity. The trained model can then be executed rapidly on low-computation modules.


\section{Conclusions and Prospects}\label{sec:conclusion}

In this work, we presented a fully integrated multiagent UAV system for searching, collecting and transporting objects with unknown locations in an outdoor environment. The proposed system simplifies such complex tasks by introducing full autonomy which extends its application domains to real life situations such as search and rescue missions and commercial package delivery. Objects were localized based on a monocular camera and the drone altitude, and picked up using our customized novel passive grasping mechanism with feedback. The overall system architecture was implemented and tested successfully in an outdoor environment using a simple yet effective approach with low-cost hardware, which makes it an appealing research testbed for future multiagent control algorithms. Further enhancements can be made in the design as well as the cooperative control techniques to incorporate robust performance of the system under varying environment conditions.


\section{Supplementary Material}
\label{sec:links}
\begin{itemize}
\item Link to video demonstration: \href{https://youtu.be/DwJcSB_iDKo}{\color{blue}{https://youtu.be/DwJcSB$\_$iDKo}}\vspace{-0.25cm}
\item Link to simulation recording: \href{https://youtu.be/fL34patISds}{\color{blue}{https://youtu.be/fL34patISds}}\vspace{-0.25cm}
\item Link to code repository: \href{https://github.com/usman094/ch-1-3}{\color{blue}{https://github.com/usman094/ch-1-3}}
\end{itemize}


\bibliography{references}

\end{document}